\documentclass[journal]{IEEEtran}
\usepackage{amsmath,amsfonts}
\usepackage{algorithmic}
\usepackage{algorithm}
\usepackage{array}
\usepackage[caption=false,font=normalsize,labelfont=sf,textfont=sf]{subfig}
\usepackage{textcomp}
\usepackage{stfloats}
\usepackage{hyperref}
\usepackage{verbatim}
\usepackage{bm}
\usepackage{color,xcolor}
\usepackage{graphicx}
\usepackage{url}
\usepackage{hyperref}
\usepackage{booktabs,makecell, multirow, tabularx, bbding, amssymb}
\begin{document}

\title{UAGLNet: Uncertainty-Aggregated Global-Local Fusion Network with Cooperative CNN-Transformer \\ for Building Extraction}
\author{Siyuan Yao,
        Dongxiu Liu, Taotao Li, Shengjie Li, Wenqi Ren and
        Xiaochun Cao,~\IEEEmembership{Senior Member,~IEEE.}

\thanks{Manuscript received January 5, 2025; revised March 12, 2025 and July 16, 2025; accepted December 10, 2025. This work was supported by the National Natural Science Foundation of China (No. 62402055, No. 62302053, No. U24B20175, No. 62322216 and No. 62025604), the Shenzhen Science and Technology Program (No. KQTD20221101093559018), the Open Fund of Key Laboratory of the Ministry of Education on Artificial Intelligence in Equipment (No. 2024-AAIE-KF04-01) and the Guangdong Basic and Applied Basic Research Foundation under Grant 2024A1515012360. (\emph{Corresponding author: Xiaochun Cao.})}

\thanks{S. Yao, D. Liu and S. Li are with School of Computer Science (National Pilot Software Engineering School), Beijing University Of Posts and Telecommunications, Beijing 100876, China. (email: yaosiyuan04@gmail.com; liudongxiu0504@gmail.com; lishengjie@bupt.edu.cn).}


\thanks{T. Li, W. Ren and X. Cao are with School of Cyber Science and Technology, Shenzhen Campus, Sun Yat-sen University, Shenzhen 518107, China. (email: litt93@mail.sysu.edu.cn; rwq.renwenqi@gmail.com; caoxiaochun@mail.sysu.edu.cn).}

}

\maketitle

\begin{abstract}
Building extraction from remote sensing images is a challenging task due to the complex structure variations of the buildings. Existing methods employ convolutional or self-attention blocks to capture the multi-scale features in the segmentation models, while the inherent gap of the feature pyramids and insufficient global-local feature integration leads to inaccurate, ambiguous extraction results. To address this issue, in this paper, we present an Uncertainty-Aggregated Global-Local Fusion Network (UAGLNet), which is capable to exploit high-quality global-local visual semantics under the guidance of uncertainty modeling. Specifically, we propose a novel cooperative encoder, which adopts hybrid CNN and transformer layers at different stages to capture the local and global visual semantics, respectively. An intermediate cooperative interaction block (CIB) is designed to narrow the gap between the local and global features when the network becomes deeper. Afterwards, we propose a Global-Local Fusion (GLF) module to complementarily fuse the global and local representations. Moreover, to mitigate the segmentation ambiguity in uncertain regions, we propose an Uncertainty-Aggregated Decoder (UAD) to explicitly estimate the pixel-wise uncertainty to enhance the segmentation accuracy. Extensive experiments demonstrate that our method achieves superior performance to other state-of-the-art methods. Our code is available at \href{https://github.com/Dstate/UAGLNet}{https://github.com/Dstate/UAGLNet.}

\end{abstract}


\section{Introduction}
\IEEEPARstart{B}{uilding} extraction is a fundamental research topic aiming to distinguish the buildings from high-resolution remote sensing (RS) images \cite{CBRnet}, which has been deployed into a wide range of applications including urban planning, population forecasting and Geographic Information System (GIS)  \cite{dong2013comprehensive,belgiu2014comparing,DBLP:journals/corr/abs-1802-09026}, etc. As satellite and aerial remote sensing imaging technologies advance rapidly, how to precisely and efficiently extract the buildings from aerial images has gained significant attention in the research community.

\begin{figure}
  \centering
  \includegraphics[width=3.4in]{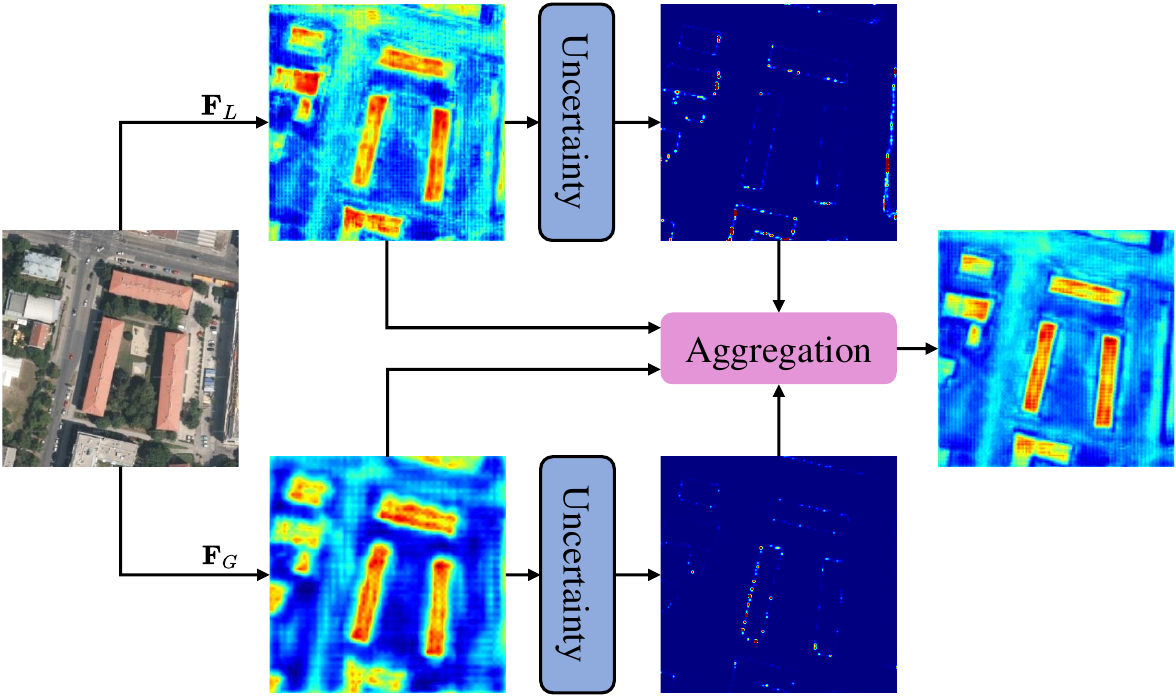}
  \caption{A general introduction to UAGLNet. We construct complementary local and global representations of the buildings under the guidance of uncertainty modeling in dual branches, which enforces the model to focus on the ambiguous regions for high-quality building extraction.}
  \label{fig:uncertainty}
\end{figure}

Over the past few decades, numerous approaches have been proposed to extract the buildings from high-resolution RS images. Early works apply hand-crafted descriptors, \emph{e.g.} texture \cite{zhang1999optimisation}, geometrical shape \cite{li2014extracting} and neighboring connection \cite{DBLP:journals/tgrs/LiFX0W15} to identify the pixels. Despite the intuitive simplicity, the manually extracted features are not effective in handling complex structure variations of the buildings. resulting in unsatisfying accuracy and limited generalization capability. Inspired by the significant successes of convolutional neural networks (CNNs), a series of CNN based building extraction methods have emerged \cite{CBRnet,DBLP:journals/tgrs/JiWL19,DBLP:journals/staeors/LiuLLLZ24,DBLP:journals/tgrs/WeiJL20}.
These methods employ local convolution kernels to capture the discriminative features via forward-backward propagation, allowing the models to construct hierarchical semantic representation of the foreground buildings by large-scale model training. Unfortunately, as the CNN based architecture relies on stacking multiple convolutional layers/blocks with limited reception fields, it fails to capture the long-range relationship in different regions. In most recent years, the vision transformers (ViTs) have been applied in building extraction task \cite{DBLP:journals/remotesensing/XuZZYL21,SDSCunet,DBLP:journals/tgrs/DingLLZCWTB22}. The ViTs based approaches utilize non-local self-attention to capture the long-range dependencies, thus being able to excavate the informative global contexts for dense prediction. The most advanced approach GraphGST \cite{DBLP:journals/tgrs/JiangSGPZSL24} aggregates the local-to-global correlations by integrating graph structural information into the Transformer architecture. The positional encoder in GraphGST learns visual semantics between adjacent pixels and uses transformer to facilitate the feature fusion. However, these methods still require to employ the hand-crafted window partition mechanisms in the transformer architecture, which are data-agnostic and ignore the input content, so the query tokens may attend to irrelevant keys/values, leading to performance degradation.

To address these issues, some works explore to utilize hybrid CNN-Transformer architecture for building extraction \cite{DBLP:journals/tgrs/WangFM022,DBLP:journals/lgrs/ZhangWZZ23,DBLP:journals/tgrs/XuLXZG23}.
For example, BuildFormer \cite{DBLP:journals/tgrs/WangFM022} designs a dual-path structure to simultaneously exploit global context and local details. DSAT-Net \cite{DBLP:journals/lgrs/ZhangWZZ23} designs a plug-and-play transformer block appended after CNN modules to supplement the global information. BCTNet \cite{DBLP:journals/tgrs/XuLXZG23} adopts two parallel branches, \emph{i.e.} a convolutional encoder branch (CB) and a transformer encoder branch (TB) to extract multi-scale feature maps. Despite the progress, the performance of existing hybrid CNN-Transformer methods are still limited by the inherent gap of the feature pyramids and insufficient integration of the global-local visual semantics. As demonstrated in Fig. \ref{fig:uncertainty}, the features in shallow layers tend to preserve more local information but are insensitive to the global contexts of the buildings. Conversely, the features in the deep layers are more favorable at exploiting the overall global connectivity of the building instance, while prone to lose the local details. Therefore, an ideal building extraction approach should not only construct high-quality global-local representation to narrow the gap of the feature pyramids, but also make full use of the complementary global-local representation for feature integration.

\begin{figure}
  \centering
  \includegraphics[width=1.0\linewidth]{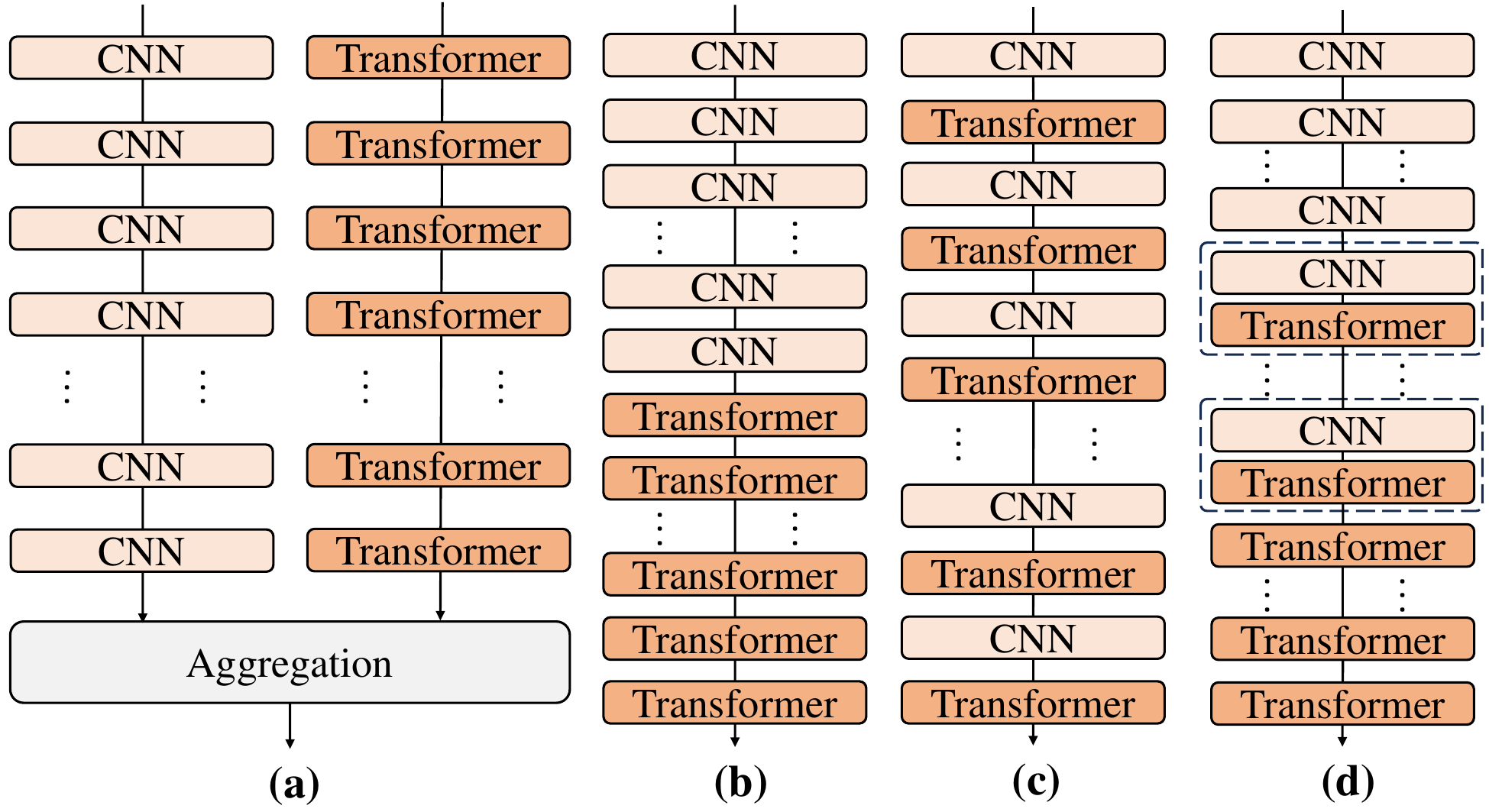}
  \caption{Comparison of different hybrid CNN-Transformer structures for building extraction. (a) Parallel CNN-Transformer structure \cite{DBLP:journals/tgrs/WangFM022}; (b) Sequentially Mixed CNN-Transformer structure by stacking CNNs in shallow layers and Transformers in deep layers \cite{DBLP:journals/corr/abs-2102-04306}; (c) Alternative CNN-Transformers structure \cite{DBLP:journals/lgrs/ZhangWZZ23}; (d) Our Cooperative architecture.}
  \label{fig:hybrid}
\end{figure}

In this paper, we propose a novel Uncertainty-Aggregated Global-Local Fusion Network (UAGLNet) to alleviate the inherent gap of the feature pyramids and insufficient global-local feature integration. As shown in Fig. \ref{fig:hybrid}, instead of stacking CNN and transformer blocks in parallel or sequential fashion like conventional hybrid CNN-Transformer models \cite{DBLP:journals/tgrs/WangFM022,DBLP:journals/corr/abs-2102-04306,DBLP:journals/lgrs/ZhangWZZ23}, 
our UAGLNet employs convolutional blocks in the early stages to capture local information, and transformer blocks in the terminate stages to capture global information. To narrow the gap of the feature pyramids, a Multi-Kernel Feature Modulator (MKFM) is designed to extract the features with different reception fields, which is combined with the standard Multi-Head Self-Attention (MHSA) in an intermediate cooperative interaction block (CIB). The features obtained by the CIB module in the third stage exhibits stronger discriminability while preserving local details. Second, to overcome the insufficient global-local feature integration, we design a Global-Local Fusion (GLF) module. This module complementarily leverages the CIB outputs to compensate the local features and global features via efficient interaction. Finally, to enforce the model's discriminative capability in challenging regions, these interacted global-local features are passed into Uncertainty Aggregated Decoder (UAD), which suppresses the low-confidence uncertain region and alleviate the segmentation uncertainties eventually.


In conclusion, our contributions can be summarized as follows:
\begin{itemize}
\item We design a Cooperative Encoder, which captures local/global information using intermediate cooperative interaction block (CIB) with hybrid convolutional and self-attention operators, allowing the model to narrow the inherent gap of the feature pyramids.
\item We propose a Global-Local Fusion (GLF) module to complementarily fuse the global and local features, making the model perform more efficient feature integration among the hierarchical pyramids.
\item We propose an Uncertainty-Aggregated Decoder (UAD), which enforces the model to focus on the challenging pixels to mitigate the segmentation uncertainty.
\end{itemize}

\section{Related Work}
\subsection{Building Extraction}

Existing building extraction methods typically employ deep learning techniques. The representative approaches utilize CNNs \cite{DBLP:journals/tgrs/JiWL19,DBLP:journals/staeors/LiuLLLZ24,DBLP:journals/remotesensing/LiuLLSYXZ19,DBLP:journals/tgrs/GuoSDZWD21} to extract the discriminative features. For example, SRINet \cite{DBLP:journals/remotesensing/LiuLLSYXZ19} aggregates multi-scale contexts for semantic understanding by successively fusing multi-level features. HRNet \cite{DBLP:conf/cvpr/0009XLW19} connects feature maps of different resolutions and performs interaction between them for semantic feature enhancement. MTPA-Net \cite{DBLP:journals/tgrs/GuoSDZWD21} proposes a multi-task parallel attention convolutional structure to improve the accuracy using the scene prior. CBR-Net \cite{CBRnet} progressively refines the building boundary by perceiving the direction of pixels to the nearest candidate object. LFEMAP-Net \cite{DBLP:journals/staeors/LiuLLLZ24} enhances the representation of spatial details under the guidance of edge prior and utilizes multi-scale attention module at the decoding stage for feature pyramid aggregation.

A key limitation of the CNN based approaches is that they fail to capture the long-range relationship of pixels in different regions. Recently, the transformer based methods have been proposed for building extraction. MSST-Net \cite{DBLP:journals/remotesensing/YuanX21} utilizes the Swin transformer as backbone and develops a scale adaptive decoder for multi-scale feature representation. STT \cite{DBLP:journals/remotesensing/ChenZS21} represents the buildings as a set of sparse feature vectors to reduce the computational complexity in transformers. BuildFormer \cite{DBLP:journals/tgrs/WangFM022} captures the global context while preserving spatial-detailed features by parallelizing convolutions and transformers. DSAT-Net \cite{DBLP:journals/lgrs/ZhangWZZ23} combines CNNs and Transformers using an efficient dual spatial attention module to maximize the advantages of them. BCT-Net \cite{DBLP:journals/tgrs/XuLXZG23} parallelizes the convolutional branch and the transformer branch to extract multi-scale feature maps. However, the aforementioned methods ignore the gap of the feature pyramids, yielding inaccurate, ambiguous extraction results due to the insufficient global-local feature integration. In this work, we make full use of the complementary global-local representations for feature integration, which significantly boosts the building extraction performance in complex scenarios.

\subsection{Hybrid CNN-Transformer Networks}
To leverage the advantage of both CNNs and transformers, the hybrid architecture has proven to be effective in computer vision tasks \cite{DBLP:conf/icml/ZhouYXXAFA22,DBLP:journals/tgrs/LiCHZQK19,DBLP:journals/tip/YaoZRMHC21,DBLP:conf/cvpr/SrinivasLPSAV21,DBLP:conf/iccv/WuXCLDY021,DBLP:conf/cvpr/Guo0WT00X22,DBLP:journals/lgrs/Wang0D0MF22,yao2025umdatrack}. CvT \cite{DBLP:conf/iccv/WuXCLDY021} first introduces convolutional neural networks before self-attention operation to construct hierarchical representation for local and spatial context modeling. CMT \cite{DBLP:conf/cvpr/Guo0WT00X22} combines depth-wise convolution and lightweight MHSA to balance the performance and computational cost between these two operators. Some works \cite{DBLP:journals/corr/abs-2102-04306,DBLP:journals/tgrs/HeZZZYX22,UnetFormer,DBLP:journals/tgrs/LiZDD22} further incorporate the transformer with classical CNN models, \emph{e.g.} UNet \cite{DBLP:conf/miccai/RonnebergerFB15} to enhance the global dependencies learning. These methods sequentially stacked CNN and transformer to capture multi-scale feature pyramids, and progressively fuse the feature pyramids using UNet decoder to recover the pixel-wise details.

Besides, the most advanced researches \cite{DBLP:journals/tgrs/WangFM022,DBLP:conf/iccv/LinWCHJ23,DBLP:journals/tip/ZhangCLCFCZK21,DBLP:journals/tgrs/ZhangCWL23,DBLP:journals/tgrs/LiuYXX23,DBLP:journals/tgrs/LiuYXZC24,DBLP:journals/tip/YaoSXWC24,DBLP:journals/tgrs/NiLCWL24,tian2025wb} point out that the parallel or asymmetric structures can provide more valuable global-local semantics in different scales. For example, ACAHNet \cite{DBLP:journals/tgrs/ZhangCWL23} combines CNN and transformer in a series-parallel manner and enhances the interaction between features by introducing an adaptively updated semantic map. SMT \cite{DBLP:conf/iccv/LinWCHJ23} proposes a scale-aware modulation mechanism to fuse the multi-scale information and expand the receptive field. TCNet \cite{xiang2024tcnet} introduces Interactive Self-attention (ISa) and Windowed Self-attention Gating (WSaG) mechanism to fuse the multi-level features. Zhang \emph{et~al.} \cite{DBLP:journals/lgrs/ZhangWZZ23} propose a dual spatial attention, where a global attention path (GAP) and a local attention path (LAP) are designed for global dependencies and local details prediction. Despite the promising successes, these methods neglect the inherent gaps of the feature pyramids and lack efficient interactions between the global and local features, which limit the segmentation accuracy in the ambiguous region.

\subsection{Uncertainty Estimation}
Uncertainty property \cite{DBLP:conf/nips/KendallG17,DBLP:conf/nips/Lakshminarayanan17,DBLP:conf/iccv/ChoiCKL19,DBLP:conf/cvpr/Wang0GFW21,DBLP:conf/cvpr/WangZWLL22,DBLP:journals/tip/YaoGYRC25} has attracted increasing attention in recent safety-critical deep models, which provides a powerful evaluator to measure the confidence and assists to identify the ambiguous predictions. There are mainly two types of uncertainty in machine learning, \emph{i.e.} aleatoric uncertainty and epistemic uncertainty. Aleatoric uncertainty is relative to the inherent variability or randomness of the data, while epistemic uncertainty reflects the model inadequacy \cite{DBLP:conf/nips/KendallG17}. For aleatoric uncertainty modeling, Gaussian YOLOv3 \cite{DBLP:conf/iccv/ChoiCKL19} incorporates a Gaussian mixture model to alleviate localization ambiguity for target objects in challenging scenarios. In \cite{DBLP:conf/nips/Lakshminarayanan17}, Lakshminarayanan \emph{et~al.} employ an adversarial perturbation technique to generate additional data to train a probabilistic neural network to model the aleatoric uncertainty. Wang \emph{et~al.} \cite{DBLP:conf/cvpr/Wang0GFW21} propose a domain adaptive segmentation network under the guidance of uncertain pseudo labels for model transfer training. UMNet \cite{DBLP:conf/cvpr/WangZWLL22} introduces a multi-source uncertainty mining method to generate reliable pixel-wise labels from multiple pseudo labels, which is capable to improve the robustness in an unsupervised learning manner.

For epistemic uncertainty, recent advances utilize probabilistic representation models that learn the distribution over network weights or features \cite{DBLP:conf/icml/BlundellCKW15}. Kendall \emph{et~al.} \cite{DBLP:conf/bmvc/KendallBC17} present a Bayesian convolutional neural networks to produce probabilistic pixel-wise semantic segmentation. Huang \emph{et~al.} \cite{DBLP:conf/eccv/HuangHCWS18} use the temporal information to simulate the sampling procedure. Postels \emph{et~al.} \cite{DBLP:conf/iccv/PostelsFCNT19} present a sampling-free approach, which approximates the epistemic uncertainty estimates using variance propagation technique. UGTR \cite{DBLP:conf/iccv/0054Z00L0F21} utilizes probabilistic information to locate difficult-to-detect regions and reason over these uncertain regions with additional attention module. In this work, we focus on identifying the challenging pixels, aiming to mitigate the segmentation ambiguity in uncertain regions.
\section{Methodology}
\subsection{Overview}
\begin{figure*}
  \centering
  \includegraphics[width=1.0\linewidth]{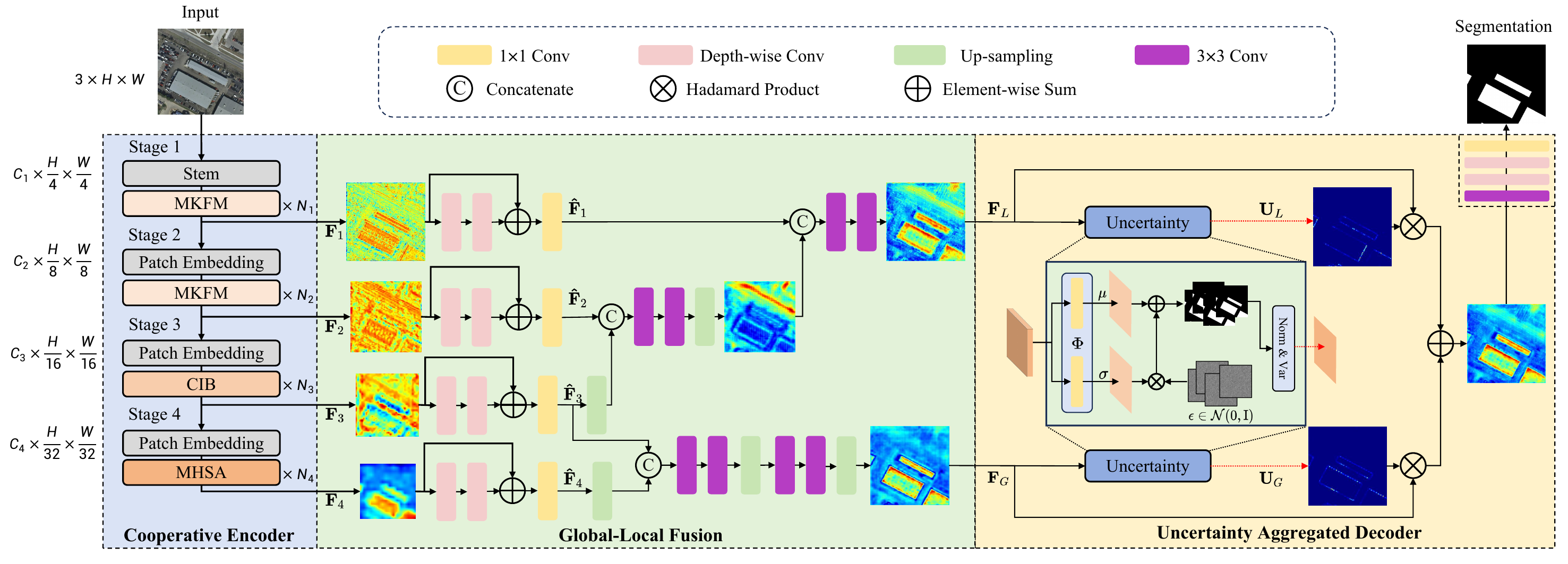}
  \caption{The overall framework of UAGLNet. UAGLNet consists of a Cooperative Encoder (CE), a Global-Local Fusion module (GLF) and an Uncertainty-Aggregated Decoder (UAD). In CE, the intermediate cooperative interaction block (CIB) is introduced to narrow the gap between the local and global features. The hierarchical features are fed into GLF and UAD for feature fusion and uncertainty aggregation.}
  \label{fig:architecture}
\end{figure*}

We present the overall structure of the proposed UAGLNet in Fig. \ref{fig:architecture}. UAGLNet follows an encoder-decoder pipeline, which consists of a Cooperative Encoder (CE), a Global-Local Fusion module (GLF) and an Uncertainty-Aggregated Decoder (UAD). The CE employs convolutional blocks in the early stages and transformer blocks in the terminate stage to capture the local and global visual semantics. Among them, an intermediate cooperative interaction block (CIB) is introduced to narrow the gap between the local and global features. Afterwards, the output features are then fed into GLF to complementarily fuse the hierarchical visual representations. Subsequently, we utilize UAD to quantify the pixel-wise uncertainty, aiming to enforce the model to focus on the unconfident regions to mitigate the segmentation uncertainty.

\subsection{Cooperative Encoder}

We propose a cooperative encoder to capture both local and global information using hybrid convolutional and self-attention operators. In the early stages, we employ hierarchical CNN blocks to construct local representation of the buildings. Then we introduce cooperative interaction block using hybrid CNN/transformer layers at the intermediate stage to enforce the global-local information interaction. For the terminate stage, we utilize a pure transformer to enhance the global contexts.

Formally, given an input image $\mathbf{I}\in \mathbb{R}^{3 \times H\times W}$, we pass it into the cooperative encoder to generate hierarchical feature maps $\mathcal{F} = \left \{ \mathbf{F}_{1},\mathbf{F}_{2},\mathbf{F}_{3},\mathbf{F}_{4} \right \}$, where $\mathbf{F}_{i}\in\mathbb{R}^{C_{i} \times {H}_{i}\times {W}_{i}}$ denotes the output feature map in the $i$-th stage. At the first two stages, we utilize CNNs with different convolutional kernels to capture the local information. The image $\mathbf{I}$ is firstly flattened and linearly projected to be a 1-dim vector, which can be reshaped as an embedding feature map $\mathbf{Z}\in\mathbb{R}^{C_{1}\times {H}_{1}\times {W}_{1}}$. Then we split the feature map $\mathbf{Z}$ across the channel dimension to construct versatile $n$ groups, \emph{i.e.} $\mathbf{Z}=[ \mathbf{Z}_{1},\mathbf{Z}_{2},\mathbf{Z}_{3},...,\mathbf{Z}_{n} ]$. For each feature group, we apply depth-wise separable convolution with different kernel sizes to capture the local information, and perform point-wise convolution to combine these grouped features, which can be given by:

\begin{equation}
\begin{aligned}
\mathbf{Z}{'} &= \mathrm{Cat}(\mathrm{DW}_{3\times3}(\mathbf{Z}_{1}),...,\mathrm{DW}_{k\times k}(\mathbf{Z}_{n}))\\
\mathbf{M} &= \mathbf{W}_{p}\ast \mathbf{Z}{'},
\end{aligned}
\end{equation}
where $\mathbf{W}_{p}$ denotes the point-wise convolution, $\mathrm{Cat}(\cdot)$ denotes the concatenation operations, $\mathrm{DW}_{k\times k}(\cdot)$ means the depth-wise separable convolutions with kernel size of $k$. The Depth-wise Convolutions is used to enhance the computationally efficient and exploit the visual semantic without introducing the channel-wise redundancy. The output feature map $\mathbf{M} \in \mathbb{R}^{C_{i} \times {H}_{i}\times {W}_{i}}$ can be regarded as a multi-kernel feature modulator (MKFM), which enhances the local representation with different reception fields. The modulation process on feature $\mathbf{F}_{i}$ can be formulated as:
\begin{equation}
\mathrm{MKFM}(\mathbf{F}_{i}) = \mathbf{M} \otimes \phi(\mathbf{F}_{i}),
\end{equation}
where $\otimes$ is the Hadamard product, $\phi(\cdot)$ is a linear embedding function with reshaped dimension. By stacking multiple MKFMs in the first two stages, UAGLNet can capture diverse multi-scale local features with only a small amount of computation costs.

\begin{figure}
  \centering
  \includegraphics[width=2.8in]{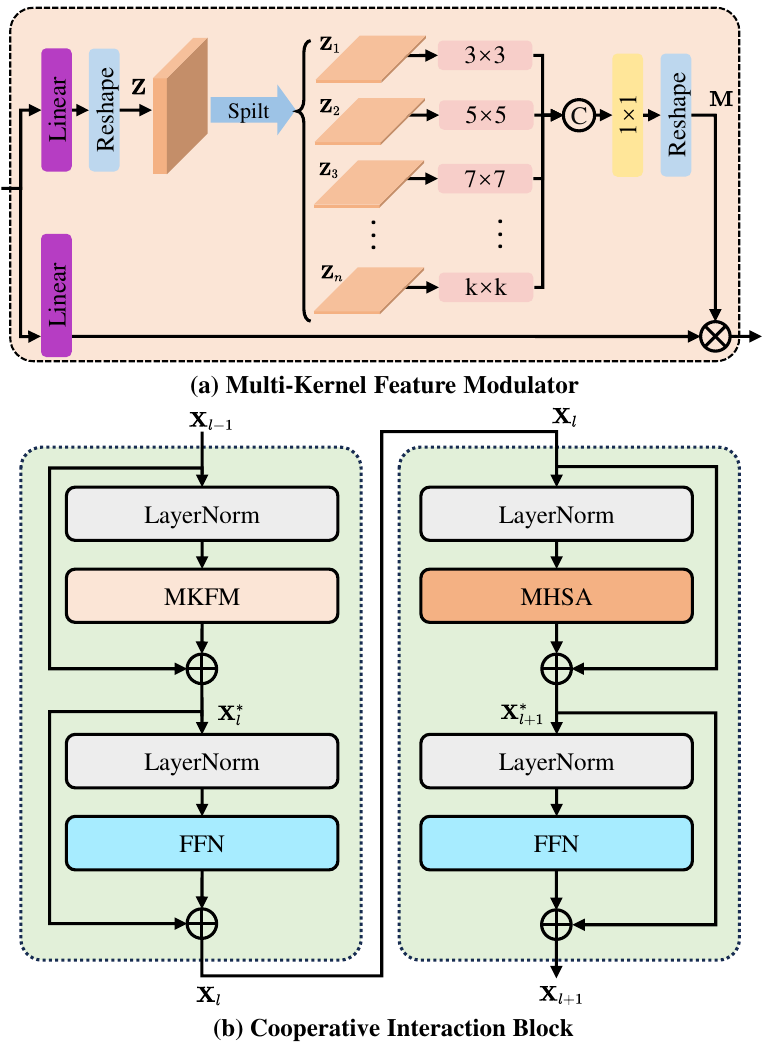}
  \caption{(a) Details of the multi-kernel feature modulator in our cooperative encoder (CE). (b) Illustration of the CIB, which stacks hybrid CNN and transformer layers for global-local information interaction.}
  \label{fig:Alter&CNN}
\end{figure}

On the contrary, the features learned in the deep stages tend to focus on the discriminative global contexts. To narrow the gap between the local and global features, we develop a cooperative interaction block (CIB) using hybrid CNN-transformer layers in the following stage. As illustrated in Fig. \ref{fig:Alter&CNN} (b), at the third stage, we stack multiple CIBs to conduct the global-local information interaction. For the $l$-th CIB in the third stage, the input feature $\mathbf{X}_{l - 1}\in \mathbb{R}^{C_{i} \times {H}_{i}\times {W}_{i}}$ is firstly normalized and fed into MKFM with residual connection as follows:

\begin{equation}
\begin{aligned}
& \mathbf{X}_{l}^{*}  = \mathbf{X}_{l-1} + \mathrm{MKFM}(\mathrm{Norm}(\mathbf{X}_{l-1})),\\
& \mathbf{X}_{l} = \mathbf{X}_{l}^{*} + \mathrm{FFN}(\mathrm{Norm}(\mathbf{X}_{l}^{*})),
\end{aligned}
\end{equation}
where $\mathbf{X}_{l}^{*}\in \mathbb{R}^{C_{i} \times {H}_{i}\times {W}_{i}}$ denotes the intermediate feature in the $l$-th block. $\mathrm{FFN}$ is the feed-forward network. Afterwards, we send $\mathbf{X}_{l}$ into a standard Multi-Head Self-Attention (MHSA) \cite{DBLP:conf/nips/VaswaniSPUJGKP17} to model the global dependencies:

\begin{equation}
\begin{aligned}
& \mathbf{X}_{l+1}^{*}  = \mathbf{X}_{l} + \mathrm{MHSA}(\mathrm{Norm}(\mathbf{X}_{l})),\\
& \mathbf{X}_{l+1} = \mathbf{X}_{l+1}^{*} + \mathrm{FFN}(\mathrm{Norm}(\mathbf{X}_{l+1}^{*})).
\end{aligned}
\end{equation}

With alternative communications between MKFM and MHSA using multiple CIBs, the cooperative encoder allows to perform effective interaction among the global and local visual semantics, thus the semantic gaps can be narrowed. Finally, in the last stage, we utilize multiple transformer blocks with MHSA to enhance the global context learning, which can be formulated by:
\begin{equation}
\mathrm{Attention}(Q,K,V) = \mathrm{Softmax}(\frac{QK^{T}}{\sqrt{C_{i}}})V,
\end{equation}
where $Q, K, V \in \mathbb{R}^{N_{i} \times C_{i}}$. $N_{i}={H}_{i}\times {W}_{i}$  represents the number of tokens and $C_{i}$ is the feature dimension. The MHSA enhances vanilla self-attention by projecting the input vector into multiple heads to perform self-attention independently to capture various aspects of the input data. The MHSA can be formulated as follows:
\begin{equation}
\begin{aligned}
\mathrm{MHSA}&(X) = \mathrm{Concat}[\mathrm{head}_{1}, \mathrm{head}_{2}, ... , \mathrm{head}_{h}]W^{O}, \\
& \mathrm{head}_{j} = \mathrm{Attention}(Q_{j},K_{j},V_{j}), \\
\end{aligned}
\end{equation}
where $X \in \mathbb{R}^{N_{i} \times C_{i}}$ is the input feature, $h$ is the number of heads, $W^{O} \in \mathbb{R}^{C_{i}\times C_{i}}$ denotes weights metrics of the linear projection layer. $Q_{j}, K_{j}, V_{j}$ represent the query, key, and value for the $j$-th head, respectively.

\subsection{Global-Local Fusion}
After obtaining the cooperative representations of the RS image, we propose a global-local fusion (GLF) module to complementarily fuse the hierarchical visual semantics. The hierarchical feature maps $\mathcal{F} = \left \{ \mathbf{F}_{1},\mathbf{F}_{2},\mathbf{F}_{3},\mathbf{F}_{4} \right \}$ are firstly passed into a depth convolutional block with residual connection to establish informative propagation pathways, which can be formulated as:
\begin{equation}
\hat{\mathbf{F}}_{i} = \mathbf{F}_{i} + \mathrm{DWConv}(\mathbf{F}_{i}),
\end{equation}
where $\mathrm{DWConv}(\cdot)$ denotes the depth convolutional block consists of two depth-wise separable convolution layers and a point-wise convolution layer.




Afterwards, we conduct feature fusion on the feature maps $\hat{\mathcal{F}} =\{\hat{\mathbf{F}}_{1}, \hat{\mathbf{F}}_{2},\hat{\mathbf{F}}_{3},\hat{\mathbf{F}}_{4} \}$. Note that the cooperative interaction block (CIB) using hybrid CNN-transformer structure is only available in the third stage, which excavates both local and global visual semantics by cooperatively merging the multi-scale local features using MHSA. Therefore, we selectively fuse $\{ \mathbf{F}_{1} , \mathbf{F}_{2} , \mathbf{F}_{3} \}$ and $\{ \mathbf{F}_{3}, \mathbf{F}_{4} \}$ to construct local and global representations. For the local representation, we utilize the CNN and  cooperative features, \emph{i.e.} $\{ \mathbf{F}_{\mathrm{1}}, \mathbf{F}_{\mathrm{2}}, \mathbf{F}_{\mathrm{3}} \}$ to enhance the local information as follows:
\begin{equation}
\label{eq:local_feature}
\mathbf{F}_{L} = \mathrm{Conv}(\mathrm{Cat}(\hat{\mathbf{F}}_{1}, \mathrm{UpConv}(\mathrm{Cat}(\hat{\mathbf{F}}_{2}, \mathrm{UpConv}(\hat{\mathbf{F}}_{3}))))),
\end{equation}
where $\mathrm{UpConv}(\cdot)$ denotes a series of convolutions with spatial up-sampling, $\mathrm{Cat}(\cdot)$ denotes the concatenation operations.

Similarly, for the global representation, we fuse features in the last two stages to establish global representation. The output $\mathbf{F}_{G}$ can be given by:
\begin{equation}
\label{eq:global_feature}
\mathbf{F}_{G} = \mathrm{UpConv}(\mathrm{Cat}(\hat{\mathbf{F}}_{3}, \mathrm{UpConv}(\hat{\mathbf{F}}_{4}))).
\end{equation}

Based on such fusion strategy, the intermediate feature $\hat{\mathbf{F}}_{3}$ generated by the CIBs can be complementarily integrated into the local and global representations, respectively.


\subsection{Uncertainty-Aggregated Decoder}
To further enhance the segmentation accuracy in the ambiguous regions, we propose an Uncertainty-Aggregated Decoder (UAD) to identify the challenging pixels. Take the local representation branch as an example, we treat the category $\mathbf{x}_{p}$ of each pixel $p$ as a random variable to predict the pixel-wise uncertainty. Formally, suppose the size of $\mathbf{F}_{L}$ is $c \times h \times w$, we pass them through two embedding functions $\Phi_{\mu}(\cdot)$ and $\Phi_{\sigma}(\cdot)$ to generate the mean map $\bm{\mu} \in \mathbb{R}^{1 \times h \times w}$ and the variance map $\bm{\sigma} \in \mathbb{R}^{1 \times h \times w}$. Afterwards, we randomly draw a sample $\epsilon$ from the Gaussian distribution $\epsilon \sim \mathcal{N}(0,\mathbf{I})$, and generate the sample by computing $\bm{\mu}+\epsilon\bm{\sigma}$. By doing this, we can obtain $T$ sampled feature maps $\mathbf{x}^{(1)}, \mathbf{x}^{(2)}, ... ,\mathbf{x}^{(T)}$ via the learned distribution. We measure the uncertainty map $\mathbf{U}$ by computing the normalized variance, which can be formulated as:
\begin{equation}
\label{eq:uncertainty}
\begin{aligned}
\bm{\mu} = \Phi_{\mu}&(\mathbf{F}_{L}) \,,\, \bm{\sigma} = \Phi_{\sigma}(\mathbf{F}_{L}) \\
\mathbf{x}_{p} &\sim \mathcal{N}(\bm{\mu}_{p},\bm{\sigma}_{p}^2),
\end{aligned}
\end{equation}
where $\mathbf{F}_{L}$ denotes the local feature, $\mathcal{N}(\bm{\mu}_{p},\bm{\sigma}_{p}^2)$ denotes the corresponding Gaussian distribution for pixel $p$. Note that the variance map $\bm{\sigma}$ also measures the pixel-wise uncertainty. If the response of arbitrary pixel in $\bm{\sigma}$ is high, it means the model is unconfident about the prediction output at this position.

Afterwards, we randomly draw $T$ samples from the probabilistic representation to generate a series of uncertainty-aware segmentation maps. We consider the turbulence of the segmentation outputs in Eq. \ref{eq:uncertainty} and calculate their variance as the uncertainty map $\mathbf{U}$, which can be formulated as:
\begin{equation}
\label{eq:uncertainty_map}
\mathbf{U} = \mathrm{Norm}(\mathrm{Var}(\mathbf{x}^{(1)}, \mathbf{x}^{(2)}, ... ,\mathbf{x}^{(T)})),
\end{equation}
where $\mathbf{x}^{(1)}, \mathbf{x}^{(2)}, ... ,\mathbf{x}^{(T)}$ means the $T$ samples, $\mathrm{Var}(\cdot)$ denotes the variance function and $\mathrm{Norm}(\cdot)$ denotes the min-max normalization.

Although intuitively feasible for uncertainty modeling, the sampling operation in Eq. \ref{eq:uncertainty_map} requires non-differentiable sampling operation to estimate the uncertainty map, making it challenging to train the UAD using forward-backward propagation. To address this issue, we adopt the reparameterization trick \cite{DBLP:journals/corr/KingmaW13} to convert the direct sampling operation to be trainable components. Specifically, instead of directly drawing samples from the probabilistic representation, we introduce the turbulence $\bm{\epsilon} \in \mathbb{R}^{1 \times h \times w}$ following the standard Gaussian distribution $\mathcal{N}(0,\mathbf{I})$ and generate $\mathbf{x} = \bm{\sigma} \times \bm{\epsilon} + \bm{\mu}$. By doing this, the non-differentiable sampling process can be avoided and the gradient information can be effectively backward propagated for uncertainty learning.

Through the aforementioned approach, we can obtain the local uncertainty $\mathbf{U}_{L}$ for $\mathbf{F}_{L}$ and the global uncertainty $\mathbf{U}_{G}$ for $\mathbf{F}_{G}$. To mitigate the impact of low-confidence regions, the uncertainty maps in both local and global branches can be regarded as attenuation weights, which aggregate $\mathbf{F}_{L}$ and $\mathbf{F}_{G}$ for feature integration:
\begin{equation}
\label{eq:uncertainty_whole}
\mathbf{F}_{\mathrm{out}} = (1 - \mathbf{U}_{G}) \times \mathbf{F}_{G} + (1-\mathbf{U}_{L}) \times \mathbf{F}_{L},
\end{equation}
where $\mathbf{F}_{\mathrm{out}}$ is the aggregated feature. With such integration, the low-confidence uncertain region of the global-local features can be effectively suppressed, which boosts the segmentation outputs in complex scenarios.

\begin{table}[t!]
    \begin{center}
    \caption{Detailed architecture specifications at four stages of the cooperative encoder (CE).}
    \renewcommand\arraystretch{1.4}
    \label{tab:CEdetail}
    \begin{tabular}{c|c|c}
        \hline
         & \multirow{2}{*}{\begin{tabular}{@{}c@{}}\textbf{Downsp rate} \\ \textbf{(Output size)} \end{tabular}} & \multirow{2}{*}{\textbf{Layers}} \\
         & & \\
        \hline
        \hline

        \multirow{4}{*}{\text{stage 1}} &
        \multirow{4}{*}{\begin{tabular}{@{}c@{}} 4 $\times$ \\ (128 $\times$ 128) \\ \end{tabular}} &
        Conv 3$\times$3, stride = 2, dim = 64 \\
        \cline{3-3}
        & & Conv 2$\times$2, stride = 2, dim = 64 \\
        \cline{3-3}
        & & \multirow{2}{*}{$\left[ \begin{array}{c} \text{MKFM, k = 9, n = 4} \\ \hline \text{FFN, mlp\_ratio = 4} \end{array} \right] \times$ 2}  \\
        & & \\
        \hline

        \multirow{3}{*}{\text{stage 2}} &
        \multirow{3}{*}{\begin{tabular}{@{}c@{}} 8 $\times$ \\ (64 $\times$ 64) \\ \end{tabular}} &
        Conv 3$\times$3, stride = 2, dim = 128 \\
        \cline{3-3}
        & & \multirow{2}{*}{$\left[ \begin{array}{c} \text{MKFM, k = 9, n = 4} \\ \hline \text{FFN, mlp\_ratio = 4} \end{array} \right] \times$ 2}  \\
        & & \\
        \hline

        \multirow{5}{*}{\text{stage 3}} &
        \multirow{5}{*}{\begin{tabular}{@{}c@{}} 16 $\times$ \\ (32 $\times$ 32) \\ \end{tabular}} &
        Conv 3$\times$3, stride = 2, dim = 256 \\
        \cline{3-3}
        & & \multirow{4}{*}{$\left[ \begin{array}{c} \text{MKFM, k = 9, n = 4} \\ \hline \text{FFN, mlp\_ratio = 4} \\ \hline \text{MHSA, num\_head = 8} \\ \hline \text{FFN, mlp\_ratio = 4} \end{array} \right] \times$ 4}  \\
        & & \\
        & & \\
        & & \\
        \hline

        \multirow{3}{*}{\text{stage 4}} &
        \multirow{3}{*}{\begin{tabular}{@{}c@{}} 32 $\times$ \\ (16 $\times$ 16) \\ \end{tabular}} &
        Conv 3$\times$3, stride = 2, dim = 512 \\
        \cline{3-3}
        & & \multirow{2}{*}{$\left[ \begin{array}{c} \text{MHSA, num\_head = 16} \\ \hline \text{FFN, mlp\_ratio = 2} \end{array} \right] \times$ 1}  \\
        & & \\
        \hline

    \end{tabular}
    \end{center}
\end{table}

\subsection{Loss Function}
We present the detailed online tracking framework of UAGLNet in Algorithm \ref{alg:UAGLNet}. To train the UAGLNet, we employ a joint loss function to minimize the difference between the ground truth mask and the predicted segmentation output, which can be formulated as:
\begin{equation}
\mathcal{L}_{\mathrm{seg}}(\mathbf{S}) = \mathcal{L}_{\mathrm{dice}}(\mathbf{S},\mathbf{Y}) + \mathcal{L}_{\mathrm{bce}}(\mathbf{S},\mathbf{Y}) + \gamma \mathcal{L}_{\mathrm{bce}}(|\mathbf{S}|,|\mathbf{Y}|),
\end{equation}
where $\mathbf{S}$ and $\mathbf{Y}$ denote the predicted output and ground truth, $\mathcal{L}_{\mathrm{dice}}$ is the dice loss and $\mathcal{L}_{\mathrm{bce}}$ is the binary cross-entropy loss. $|\mathbf{S}|$ and $|\mathbf{Y}|$ represent the building boundaries extracted by the Laplacian convolution \cite{DBLP:conf/cvpr/FanLHWCLW21} with a kernel of $ \left[ \begin{smallmatrix}-1 & -1 & -1\\ -1 & 8 & -1\\ -1 & -1 & -1 \end{smallmatrix} \right]$.

Besides, we also introduce an uncertainty loss $\mathcal{L}_{\mathrm{unc}}$ to supervise the segmentation uncertainty, which can be given by:
\begin{equation}
\label{eq:uncertainty_seg}
\mathcal{L}_{\mathrm{unc}}(\mathbf{U}) = \mathcal{L}_{\mathrm{bce}}(\mathbf{x}^{\ast},\mathbf{Y}) + \eta \mathcal{D}(\mathcal{N}(\bm{\mu},\bm{\sigma}^{2})\,||\,\mathcal{N}(0,\mathbf{I})),
\end{equation}
where $\mathbf{x}^{\ast}$ is a sample randomly drawn from the probabilistic distribution, $\mathbf{Y}$ is the ground truth mask, $\mathcal{D}$ is the Kullback-Leibler (KL) divergence loss, $\eta$ is a hyperparameters used to balance the weights of the loss terms. Here we consider the uncertainty in both global and local branches for joint optimization. The complete loss function $\mathcal{L}_{\mathrm{total}}$ can be formulated by:
\begin{equation}
\mathcal{L}_{\mathrm{total}} = \mathcal{L}_{\mathrm{seg}}(\mathbf{S}) + \lambda_{1} \mathcal{L}_{\mathrm{unc}}(\mathbf{U}_{G}) + \lambda_{2} \mathcal{L}_{\mathrm{unc}}(\mathbf{U}_{L}),
\end{equation}
where $\lambda_{1}$ and $\lambda_{2}$ are the combination weights used to balance the influence of objectives.

\begin{figure*}[ht!]
  \centering
  \includegraphics[width=6.4in]{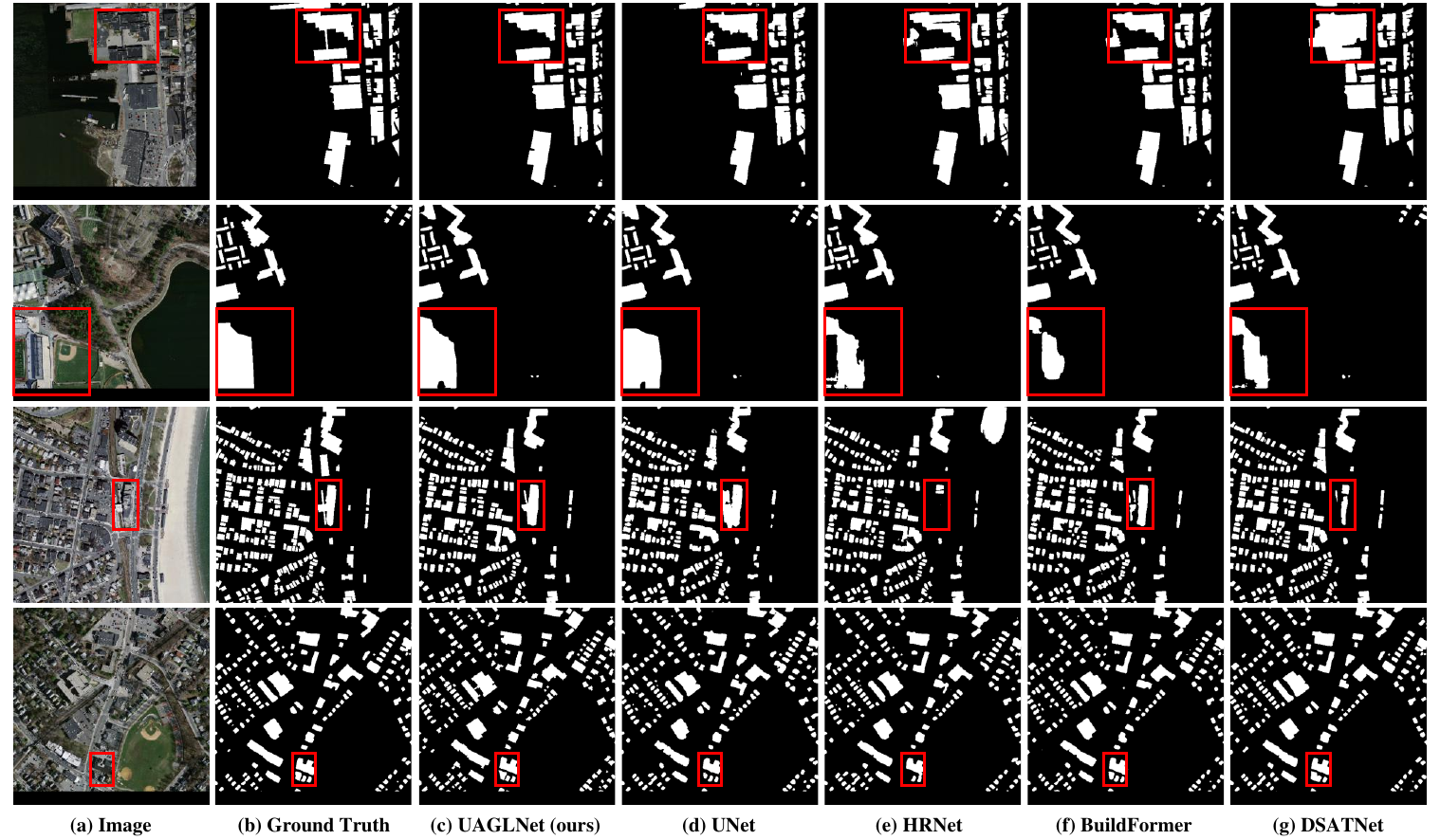}
  \caption{Visual comparison with the representative state-of-the-art methods on Massachusetts building dataset. (a) Image; (b) Ground Truth; (c) UAGLNet; (d) UNet; (e) HRNet; (f) BuildFormer; (g) DSATNet.}
  \label{fig:vis_mass}
  \vspace{-15pt}
\end{figure*}

\begin{algorithm}
    \caption{Pseudocode of the Proposed UAGLNet}
    \label{alg:UAGLNet}
    \renewcommand{\algorithmicrequire}{\textbf{Input: }}
    \renewcommand{\algorithmicensure}{\textbf{Output: }}
    \algorithmicrequire Input image $\mathbf{I}$; Ground truth mask $\mathbf{Y}$;\\
    \algorithmicensure Segmentation output $\mathbf{S}$; Local uncertainty $\mathbf{U}_{L}$; Global uncertainty $\mathbf{U}_{G}$;
    \begin{algorithmic}[1]
        \FOR{\textit{Epochs}}{
            \FOR{\textit{Batches}}{
            \STATE $\mathbf{F}_{1} \xleftarrow{\mathrm{MKFM}} \mathbf{I}$, $\mathbf{F}_{2} \xleftarrow{\mathrm{MKFM}} \mathbf{F}_{1}$, $\mathbf{F}_{3} \xleftarrow{ \; \mathrm{CIB} \; } \mathbf{F}_{2}$, $\mathbf{F}_{4} \xleftarrow{\mathrm{MHSA}} \mathbf{F}_{3}$.
            \STATE//* \textit{Global-Local Fusion via GLF} *//
            \STATE $\mathbf{F}_{L} \xleftarrow{\mathrm{GLF}} {\mathbf{F}_{1}, \mathbf{F}_{2}, \mathbf{F}_{3}}, \mathrm{and} ~\mathbf{F}_{G} \xleftarrow{\mathrm{GLF}} {\mathbf{F}_{3}, \mathbf{F}_{4}}$.
            \STATE Predict the pixel-wise uncertainty $\bm{\mu}, \bm{\sigma}$ by Eq. \ref{eq:uncertainty}.
            \STATE Randomly draw $T$ samples $\mathbf{x}^{(1)}, \mathbf{x}^{(2)}, ... ,\mathbf{x}^{(T)}$ from $\mathcal{N}(\bm{\mu},\bm{\sigma}^2)$.
            \STATE Calculate local and global uncertainty maps $\mathbf{U}_{L}$ and $\mathbf{U}_{G}$ via Eq. \ref{eq:uncertainty_map}.
            \STATE $\mathbf{F}_{out} \leftarrow (1 - \mathbf{U}_{G}) \times \mathbf{F}_{G} + (1-\mathbf{U}_{L}) \times \mathbf{F}_{L}$.
            \STATE $\mathbf{S} \leftarrow$ Squeeze the incorporated feature $\mathbf{F}_{out}$ to obtain binary prediction.
            \STATE $\mathcal{L}_{\mathrm{seg}}(\mathbf{S}) \leftarrow$ use $\mathbf{S}$ and $\mathbf{Y}$ to compute segmentation loss.
            \STATE $\mathcal{L}_{\mathrm{unc}}(\mathbf{U}_{L}),\mathcal{L}_{\mathrm{unc}}(\mathbf{U}_{G})  \leftarrow$ Compute local and global uncertainty losses using Eq. \ref{eq:uncertainty_seg}.
            \STATE $\mathcal{L}_{\mathrm{total}} \leftarrow \mathcal{L}_{\mathrm{seg}}(\mathbf{S}) + \lambda_{1} \mathcal{L}_{\mathrm{unc}}(\mathbf{U}_{G}) + \lambda_{2}\mathcal{L}_{\mathrm{unc}}(\mathbf{U}_{L})$.
            \STATE Minimize $\mathcal{L}_{\mathrm{total}}$ and update the model parameters accordingly.

            }\ENDFOR
        }\ENDFOR
    \end{algorithmic}
\end{algorithm}

\section{Experiments}
To evaluate the performance of the proposed UAGLNet, we conduct experiments on three publicly available building extraction datasets, \emph{i.e.} Inria Aerial dataset \cite{DBLP:conf/igarss/MaggioriTCA17}, Massachusetts dataset \cite{DBLP:phd/ca/Mnih13} and WHU building dataset \cite{DBLP:journals/tgrs/JiWL19}. We compare UAGLNet with multiple state-of-the-art CNN-based and Transformer-based methods, including UNet \cite{DBLP:conf/miccai/RonnebergerFB15}, SIUNet \cite{DBLP:journals/tgrs/JiWL19}, HRNet \cite{DBLP:conf/cvpr/0009XLW19}, DSNet \cite{DBLP:journals/tnn/ZhangLYYZ22}, SwinUNet \cite{DBLP:conf/eccv/CaoWCJZTW22}, TransFuse \cite{DBLP:conf/miccai/ZhangLH21}, STT \cite{DBLP:journals/remotesensing/ChenZS21}, DMBCNet \cite{DBLP:journals/remotesensing/ShiZ21}, BOMSNet \cite{DBLP:journals/tgrs/ZhouCWLLXM22}, LCS \cite{DBLP:journals/tgrs/LiuSO22}, CBRNet \cite{CBRnet}, BuildFormer \cite{DBLP:journals/tgrs/WangFM022}, DSATNet \cite{DBLP:journals/lgrs/ZhangWZZ23}, SDSCUNet \cite{SDSCunet}, UANet \cite{DBLP:journals/tgrs/LiHCZZ24}, MAFCN \cite{DBLP:journals/tgrs/WeiJL20}, BRRNet \cite{DBLP:journals/remotesensing/ShaoTWSYS20}, MSNet \cite{DBLP:journals/tgrs/LiuZZH22}, BCTNet \cite{DBLP:journals/tgrs/XuLXZG23}, FDNet \cite{DBLP:journals/tgrs/GuoSWDZ23}, MAPNet \cite{DBLP:journals/tgrs/ZhuLHML21} and LFEMAPNet \cite{DBLP:journals/staeors/LiuLLLZ24}.

\subsection{Implementation and Evaluation Details}

\noindent\textbf{Implementation Setting.}  We run the proposed UAGLNet using PyTorch framework on a single NVIDIA RTX A6000 GPU with 48GB memory. The detailed architecture of the cooperative encoder is presented in Tab. \ref{tab:CEdetail}. The remote sensing images in all datasets are cropped to $512\times512$ for training and testing phase. The data augmentation operations including random crop, random horizontal flipping, photometric distortion and mixup are used to improve the model's robustness. During training, the AdamW optimizer is employed to train the models. The learning rate is set to $5 \times 10^{-4}$ with the weight decay $0.01$. To mitigate the overfitting problem. we utilize residual connections in CIB and MKFM modules to facilitate the gradient propagation, enabling the network to learn robust features. We also employ Cosine Annealing Warm Restarts as our learning rate scheduler to prevent the model from getting stuck in local minima. The hyper-parameter $\lambda_{1}$ and $\lambda_{2}$ are both $0.5$ and the scaling parameter $\eta$ is set to $0.2$. The model is trained with $105$ epochs. The batch size is set to $16$ for Inria dataset and $8$ for Massachusetts and WHU datasets. On the Inria Aerial dataset and the WHU dataset, the random drop path technique rate is set to $0.2$, while on the Massachusetts Dataset with only 151 aerial images, it is increased to $0.4$.

\noindent\textbf{Evaluation Metrics.} To evaluate the performance of different methods, we adopt the commonly used evaluation metrics in remote sensing datasets for fair comparison. The Precision ($\mathrm{P}$), Recall ($\mathrm{R}$), $\mathrm{F1}$ score and Intersection over Union ($\mathrm{IoU}$) are employed to evaluate the performance. These metrics are calculated by:

\begin{equation}
\begin{aligned}
    \mathrm{P} &= \frac{\mathrm{TP}}{\mathrm{TP}+\mathrm{FP}}\\
    \mathrm{R} &= \frac{\mathrm{TP}}{\mathrm{TP}+\mathrm{FN}}\\
    \mathrm{F1} &= \frac{2 \times \mathrm{P} \times \mathrm{R}}{\mathrm{P} + \mathrm{R}}\\
    \mathrm{IoU} &= \frac{\mathrm{TP}}{\mathrm{TP}+\mathrm{FN}+\mathrm{FP}},
\end{aligned}
\end{equation}
where $\mathrm{TP}$, $\mathrm{FP}$, $\mathrm{TN}$ and $\mathrm{FN}$ represent the pixel-wise predictions of true positive, false positive, true negative, and false negative, respectively.

\begin{table}
  \begin{center}
    \caption{Comparison Of The State-Of-The-Art Methods And Ours On Massachusetts Building Dataset}
    \label{Comparison:Mass}
    \renewcommand\arraystretch{1.2}
    \begin{tabular}{c|c| c c c c }
    \Xhline{1pt}
        Method &  Year & IoU(\%) & F1(\%) & P(\%) & R(\%) \\ \hline
        UNet \cite{DBLP:conf/miccai/RonnebergerFB15} & 2015 & 72.59 & 84.12 & 84.46 & 83.77 \\
        HRNet \cite{DBLP:conf/cvpr/0009XLW19} & 2019 & 72.89 & 84.32 & 85.83 & 82.86 \\
        MAFCN \cite{DBLP:journals/tgrs/WeiJL20} & 2020 & 73.80 & 84.93 & 87.07 & 82.89 \\
        BRRNet \cite{DBLP:journals/remotesensing/ShaoTWSYS20} & 2020 & 73.25 & 84.56 & - & - \\
        DSNet \cite{DBLP:journals/tnn/ZhangLYYZ22} & 2020 & 75.04 & 85.74 & 87.56 & 83.99 \\
        STT \cite{DBLP:journals/remotesensing/ChenZS21} & 2021 & 74.51 & 85.40 & 86.55 & 84.27 \\
        BOMSNet \cite{DBLP:journals/tgrs/ZhouCWLLXM22} & 2022 & 74.71 & 85.13 & 86.64 & 83.68 \\
        MSNet \cite{DBLP:journals/tgrs/LiuZZH22} & 2022 & 70.21 & 79.33 & 78.54 & 80.14 \\
        CBRNet \cite{CBRnet} & 2022 & 74.55 & 85.42 & 86.50 & 84.36 \\
        BuildFormer \cite{DBLP:journals/tgrs/WangFM022} & 2022 & 75.74 & 86.19 & 87.52 & 84.90 \\
        DSATNet \cite{DBLP:journals/lgrs/ZhangWZZ23} & 2023 & 76.54 & 86.71 & 87.94 & 85.52 \\
        SDSCUNet \cite{SDSCunet} & 2023 & 76.71 & 86.82 & 88.05 & 85.62 \\
        BCTNet \cite{DBLP:journals/tgrs/XuLXZG23} & 2023 & 75.04 & 85.74 & 87.95 & 83.02 \\
        FDNet \cite{DBLP:journals/tgrs/GuoSWDZ23} & 2023 & 74.54 & 85.42 & 87.95 & 83.02 \\
        UANet \cite{DBLP:journals/tgrs/LiHCZZ24} & 2024 & 76.41 & 86.63 & 87.94 & 85.35 \\ \hline
        UAGLNet (ours) & & \textbf{76.97} & \textbf{86.99} & \textbf{88.28} & \textbf{85.73} \\ 
    \Xhline{1pt}
    \end{tabular}
    \vspace{-10pt}
  \end{center}
\end{table}

\begin{figure*}[ht!]
  \centering
  \includegraphics[width=6.4in]{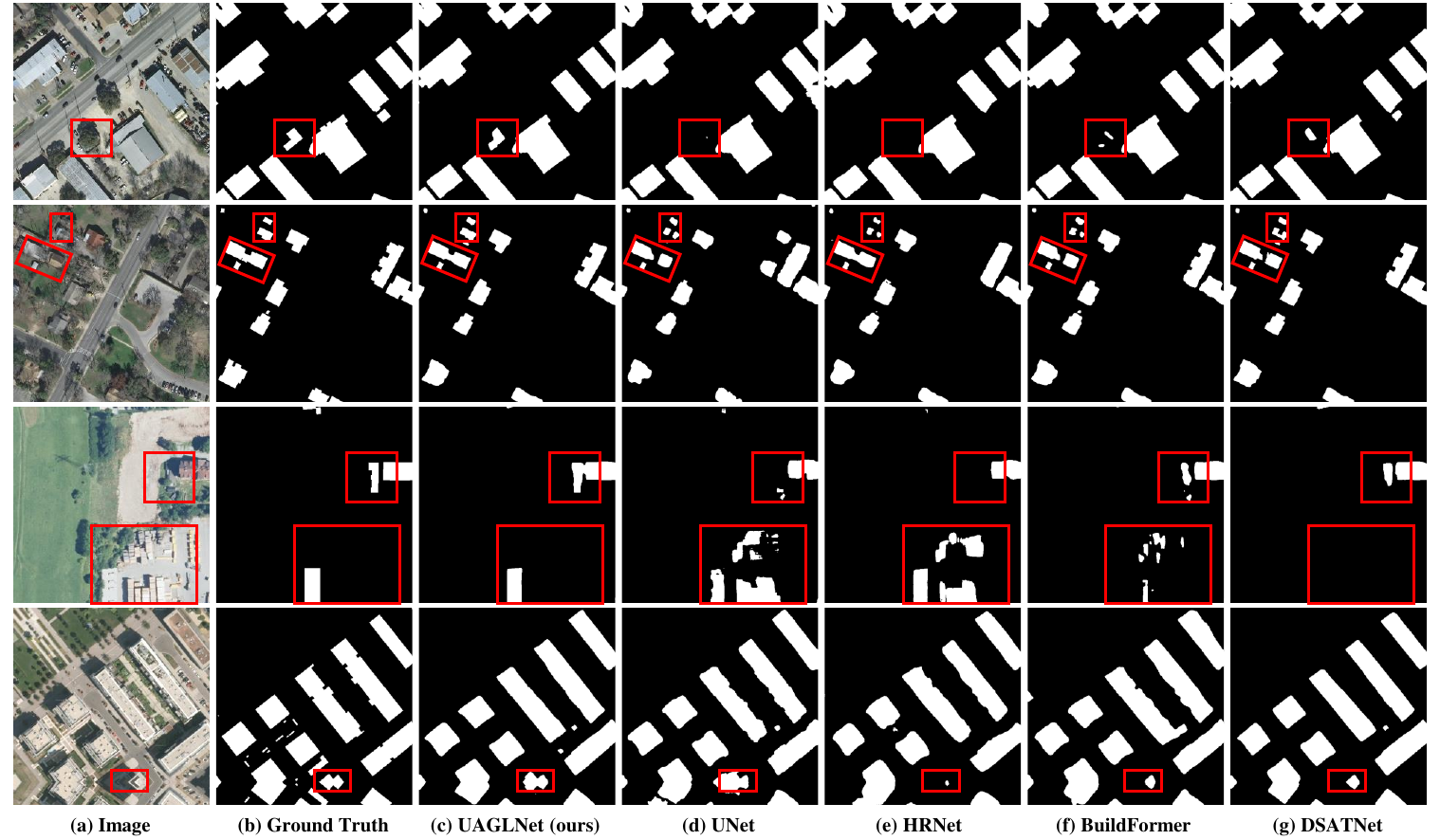}
  \caption{Visual comparison with the representative state-of-the-art methods on Inria Aerial Image Labeling dataset. (a) Image; (b) Ground Truth; (c) UAGLNet; (d) UNet; (e) HRNet; (f) BuildFormer; (g) DSATNet.}
  \label{fig:vis_inria}
  \vspace{-10pt}
\end{figure*}

\subsection{Massachusetts Dataset}
\noindent\textbf{Massachusetts Dataset.} The Massachusetts dataset \footnote{\url{https://www.cs.toronto.edu/~vmnih/data/}} contains $151$ aerial images of the Boston area with the size of $1500 \times 1500$ pixels and a spatial resolution of $1$ m/pixel. The dataset is divided into a training set, a validation set, and a test set with $137$, $4$, and $10$ images, respectively. Each image with the size of $1500 \times 1500$ is annotated by pixel-wise labels. The spatial resolution of the whole dataset is $1.2$m per-pixel, covering around $340$ km$^{2}$.

\noindent\textbf{Evaluation Results.} Table \ref{Comparison:Mass} shows the quantitative comparison of various methods on the Massachusetts building dataset. UAGLNet achieves the best performance with the precision $88.28\%$, the recall $85.73\%$ and the IoU $76.97\%$. Compared to BuildFormer, our UAGLNet obtains performance gains of $1.23\%$ in IoU and $0.76\%$ in precision. For the most recent state-of-the-art method UANet \cite{DBLP:journals/tgrs/LiHCZZ24}, UAGLNet surpasses it by $0.56\%$ in the IoU metric, $0.36\%$ in the F1 metric, $0.34\%$ in the precision metric, and $0.38\%$ in the recall metric, respectively. The superior performance on this challenging dataset confirms the effectiveness of the proposed method.

We further demonstrate the segmentation results in Fig. \ref{fig:vis_mass}, note that the Massachusetts dataset contains large quantities of small buildings with low resolutions, The global-local fusion (GLF) and uncertainty-aggregated decoder (UAD) in UAGLNet can capture high-quality local details and mitigate the ambiguity in the neighboring regions, thus the extraction results are better than HRNet, BuildFormer and DSATNet.

\begin{table*}[ht!]
    \begin{center}
    \caption{Comparison Of The State-Of-The-Art Methods And Ours On Inria Aerial Image Labeling Dataset}
    \label{Comparison:Inria}
    \renewcommand\arraystretch{1.2}
    \begin{minipage}{0.9\textwidth}
    \begin{tabular*}{\textwidth}{>{\centering}m{2.2cm}|>{\centering}m{1cm}| @{\extracolsep{\fill}} c c c c c c}
    \Xhline{1pt}
    Method & Year & IoU(\%) & F1(\%) & P(\%) & R(\%) & FLOPs & Params \\ \hline
    UNet \cite{DBLP:conf/miccai/RonnebergerFB15} & 2015 & 75.66 & 86.14 & 85.88 & 86.40 & 202.95G & 32.09M \\
    SIUNet \cite{DBLP:journals/tgrs/JiWL19} & 2019 & 71.40 & 83.33 & 84.60 & 82.10 & 215.85G & 31.18M \\
    HRNet \cite{DBLP:conf/cvpr/0009XLW19} & 2019 & 77.14 & 87.10 & 89.04 & 85.24 & 90.39G & 67.17M \\
    DSNet \cite{DBLP:journals/tnn/ZhangLYYZ22} & 2020 & 81.02 & 89.52 & 90.32 & 88.73 & - & - \\
    SwinUNet \cite{DBLP:conf/eccv/CaoWCJZTW22} & 2021 & 76.27 & 86.54 & 86.87 & 86.21 & 97.69G & 84.02M \\
    TransFuse \cite{DBLP:conf/miccai/ZhangLH21} & 2021 & 80.71 & 90.29 & 88.38 & 89.32 & 50.53G & 31.87M \\
    STT \cite{DBLP:journals/remotesensing/ChenZS21} & 2021 & 79.42 & 87.99 & - & - & 106.21G & 18.87M \\
    DMBCNet \cite{DBLP:journals/remotesensing/ShiZ21} & 2021 & 80.74 & 89.35 & 89.94 & 88.77 & 60.01G & 30.01M \\
    BOMSNet \cite{DBLP:journals/tgrs/ZhouCWLLXM22} & 2022 & 78.18 & 87.75 & 87.93 & 87.58 & - & 129.32M \\
    LCS \cite{DBLP:journals/tgrs/LiuSO22} & 2022 & 78.82 & 88.15 & 89.58 & 86.77 & 135.98G & 20.16M \\
    CBRNet \cite{CBRnet} & 2022 & 81.10 & 89.56 & 89.93 & 89.20 & 186.56G & 22.69M \\
    BuildFormer \cite{DBLP:journals/tgrs/WangFM022} & 2022 & 81.44 & 89.77 & 90.75 & 88.81 & 117.12G & 40.52M \\
    DSATNet \cite{DBLP:journals/lgrs/ZhangWZZ23} & 2023 & 82.68 & 90.52 & 91.62 & 89.44 & 57.75G & 48.50M \\
    SDSCUNet \cite{SDSCunet} & 2023 & 83.01 & 90.71 & 91.52 & 89.92 & 29.82G & 21.32M \\
    UANet \cite{DBLP:journals/tgrs/LiHCZZ24} & 2024 & 83.08 & 90.76 & 92.04 & 89.52 & 111.11G & 15.59M \\ \hline
    UAGLNet (ours) & & \textbf{83.74} & \textbf{91.15} & \textbf{92.09} & \textbf{90.22} & \textbf{28.90G} & \textbf{15.34M} \\
    \Xhline{1pt}
    \end{tabular*}
    \end{minipage}
   \end{center}
\end{table*}

\subsection{Inria Aerial Dataset}
\noindent\textbf{Inria Aerial Dataset.} Inria aerial dataset \footnote{\url{https://project.inria.fr/aerialimagelabeling/}} is a widely used urban buildings extraction dataset for semantic segmentation. There are 360 images in this dataset with the size of $5000 \times 5000$ and spatial resolution of $0.3$m, wherein $180$ are presented with the ground truth labels. These images are collected in five cities: Austin (U.S.), Chicago (U.S.), Kitsap (U.S.), Tyrol (Austria), and Vienna (Austria). Similar to \cite{DBLP:journals/tgrs/LiHCZZ24}, the original images are padded into $5120 \times 5120$ pixels and then cropped into tiles of $512 \times 512$ pixels. As a result, $9737$ and $1942$ image tiles are used for training and validation, respectively.

\noindent\textbf{Evaluation Results.} Table \ref{Comparison:Inria} shows the quantitative comparison of various methods on the Inria aerial dataset. UAGLNet achieves the best performance with the precision and recall scores of $(92.09\%, 90.22\%)$, the IoU and F1 scores of $(83.74\%, 91.15\%)$.
our UAGLNet outperforms BuildFormer by $2.30\%$ in IoU, $1.38\%$ in F1 score, $1.34\%$ in the precision, and $1.41\%$ in the recall metric. Compared to the most recent state-of-the-art method UANet \cite{DBLP:journals/tgrs/LiHCZZ24}, UAGLNet obtains the performance gains of $0.66\%$ and $0.39\%$ in the IoU and F1 metrics, respectively. Fig. \ref{fig:vis_inria} demonstrates the comparative visualization results on Inria aerial dataset. We can see that our UAGLNet also outperforms other building extraction methods in recognizing the hard building pixels and maintaining the integrity of buildings, which benefits from the cooperative global-local contexts integration and segmentation uncertainty mitigation.

Moreover, we also observe that the proposed UAGLNet achieves superior performance using the fewest parameters and floating-point operations among all methods. UGALNet only requires $28.90$G flops and $15.34$M parameters, which significantly reduce the computation and memory costs. Compared with BuildFormer, it saves $75.32\%$ computational complexity and $62.14\%$ parameters. For the efficient ViT-based U-shaped model SDSC-UNet, UGALNet also reduces $28.05\%$ parameters. This is because our UGALNet utilizes depth-wise separable convolutions and point-wise convolution in MKFMs and CIBs at the first three stages, thus the computation and memory costs are greatly decreased, allowing the model to capture the local and global information with efficiency and effectiveness.

\begin{figure*}[ht!]
  \centering
  \includegraphics[width=6.4in]{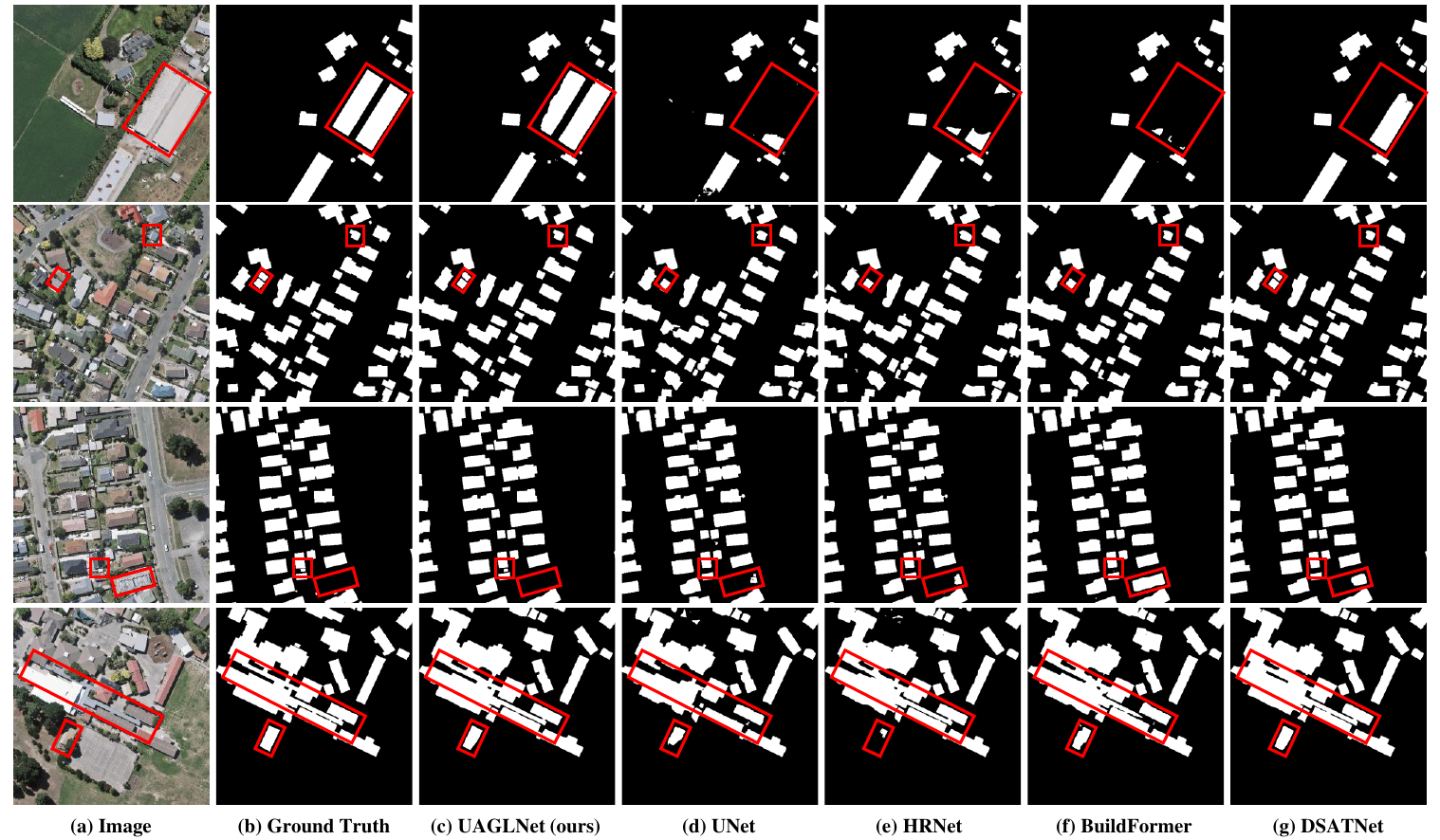}
  \caption{Visual comparison with the representative state-of-the-art methods on WHU building dataset. (a) Image; (b) Ground Truth; (c) UAGLNet; (d) UNet; (e) HRNet; (f) BuildFormer; (g) DSATNet.}
  \label{fig:vis_whu}
  \vspace{-15pt}
\end{figure*}

\begin{table}[t]
  \begin{center}
    \caption{Comparison Of The State-Of-The-Art Methods And Ours On WHU Building Dataset.}
    \renewcommand\arraystretch{1.2}
    \label{Comparison:WHU}
    \begin{tabular}{c|c|c c c c}
    \Xhline{1pt}
        Method & Year & IoU(\%) & F1(\%) & P(\%) & R(\%) \\ \hline
        UNet \cite{DBLP:conf/miccai/RonnebergerFB15} & 2015 & 88.08 & 93.66 & 93.17 & 94.16 \\
        SIUNet \cite{DBLP:journals/tgrs/JiWL19} & 2019 & 88.40 & 93.85 & 93.80 & 93.90 \\
        HRNet \cite{DBLP:conf/cvpr/0009XLW19} & 2019 & 88.21 & 93.73 & 93.63 & 93.84 \\
        MAFCN \cite{DBLP:journals/tgrs/WeiJL20} & 2020 & 90.70 & 95.04 & 95.20 & 95.10 \\
        DSNet \cite{DBLP:journals/tnn/ZhangLYYZ22} & 2020 & 89.54 & 94.48 & 94.05 & 94.91 \\
        STT \cite{DBLP:journals/remotesensing/ChenZS21} & 2021 & 90.48 & 94.97 & - & - \\
        MAPNet \cite{DBLP:journals/tgrs/ZhuLHML21} & 2021 & 90.86 & 95.21 & 95.62 & 94.81 \\
        DMBC-Net \cite{DBLP:journals/remotesensing/ShiZ21} & 2021 & 91.66 & 95.65 & 96.15 & 95.16 \\
        BOMSNet \cite{DBLP:journals/tgrs/ZhouCWLLXM22} & 2022 & 90.15 & 94.80 & 95.14 & 94.50 \\
        LCS \cite{DBLP:journals/tgrs/LiuSO22} & 2022 & 90.71 & 95.12 & 95.38 & 94.86 \\
        MSNet \cite{DBLP:journals/tgrs/LiuZZH22}  & 2022 & 89.07 & 93.96 & 94.83 & 93.12 \\
        BuildFormer \cite{DBLP:journals/tgrs/WangFM022} & 2022 & 91.44 & 95.53 & 95.65 & 95.40 \\
        BCT-Net \cite{DBLP:journals/tgrs/XuLXZG23} & 2023 & 91.15 & 95.37 & 95.47 & 95.27 \\
        FDNet \cite{DBLP:journals/tgrs/GuoSWDZ23} & 2023 & 91.14 & 95.36 & 95.27 & 95.46 \\
        LFEMAPNet \cite{DBLP:journals/staeors/LiuLLLZ24} & 2024 & 91.48 & 95.55 & 95.65 & 95.45 \\ \hline
        UAGLNet (ours) &  & \textbf{92.07} & \textbf{95.87} & \textbf{96.21} & \textbf{95.54}  \\ 
    \Xhline{1pt}
    \end{tabular}
    \vspace{-15pt}
  \end{center}
\end{table}

\subsection{WHU Dataset}
\noindent\textbf{WHU Dataset.} The WHU building dataset \footnote{\url{https://gpcv.whu.edu.cn/data/}} contains $8189$ image tiles with a spatial resolution of $0.3$ m/pixel, and the size of each image is $512 \times 512$. There are $4736$ tiles selected for training, $1036$ tiles for validation, and $2416$ tiles for testing. The whole dataset covers a huge area of over $450$ $\mathrm{km}^{2}$, including about $187000$ buildings with different scales and shapes.

\noindent\textbf{Evaluation Results.} Table \ref{Comparison:WHU} shows the quantitative comparison of various methods on the WHU building dataset. UAGLNet obtains the IoU score of $92.07\%$ and the F1 score of $95.87\%$, which outperforms recent state-of-the-art method LFEMAPNet \cite{DBLP:journals/staeors/LiuLLLZ24} by $0.59\%$ and $0.56\%$ in the IoU and precision metric, respectively. We also present the comparative visualization results in Fig. \ref{fig:vis_whu}. The UGALNet is more powerful in extracting buildings with occluded trees, and it can identify the challenging cases that closely resemble their surroundings.

\begin{table}[h]
  \begin{center}
    \caption{Ablation Experimental Results Of Variants With Different Components On Inria Aerial Image Labeling Dataset}
    \label{Ablation:Components}
    \renewcommand\arraystretch{1.4}
    \begin{tabular}{ >{\centering}m{1cm}  >{\centering\arraybackslash}m{1cm}  >{\centering\arraybackslash}m{0.8cm}  >{\centering\arraybackslash}m{0.9cm} | >{\centering\arraybackslash}m{1.4cm} >{\centering\arraybackslash}m{0.8cm}}
        \Xhline{1pt}
    CE & GLF & $ \mathbf{U}_{L}$ & $\mathbf{U}_{G}$ & IoU(\%) & F1(\%) \\ \hline
    \CheckmarkBold & & & & 82.46 & 90.39 \\
    \CheckmarkBold & \CheckmarkBold & & & 83.33 & 90.90 \\
    \CheckmarkBold & \CheckmarkBold & \CheckmarkBold & & 83.58 & 91.06 \\
    \CheckmarkBold & \CheckmarkBold & & \CheckmarkBold & 83.62 & 91.08 \\
    \CheckmarkBold & \CheckmarkBold & \CheckmarkBold & \CheckmarkBold & \textbf{83.74} & \textbf{91.15} \\ 
        \Xhline{1pt}
    \end{tabular}
    \vspace{-15pt}
  \end{center}
\end{table}

\subsection{Ablation Study}
\noindent\textbf{Ablation Study of Each Component.} To verify the effectiveness of each component in our UAGLNet, we selectively choose the cooperative encoder (CE), the global-local fusion module (GLF) and the uncertainty-aggregated decoder (UAD) for comparison. Here we classify our UAD into the local uncertainty $\mathbf{U}_{L}$ and the global uncertainty $\mathbf{U}_{G}$ to strengthen our experiments. The ablation results are shown in Table \ref{Ablation:Components}. The variant approach only with CE achieves $82.46\%$ and $90.39\%$ in the IoU and F1 metrics. By introducing GLF to generate refined global and local features, the variant approach obtains the performance improvement of $0.87\%$ and $0.51\%$, respectively. If the local uncertainty $\mathbf{U}_{L}$ or the global uncertainty $\mathbf{U}_{G}$ is further introduced, the variants can gain the IoU scores of $0.25\%$ or $0.29\%$, respectively. If the complete UAD is introduced, the variant can achieve obvious improvement of $0.41\%$. The ablation results demonstrate the effectiveness of each module in our UAGLNet.

\begin{table}[h]
  \begin{center}
    \caption{Comparison Of Hybrid CNN-Transformer Architectures On Inria Aerial Image Labeling Dataset}
    \renewcommand\arraystretch{1.4}
    \label{Comparison:Hybrid}
    \begin{tabular}{>{\centering}m{2.2cm}|>{\centering\arraybackslash}m{1cm} >{\centering\arraybackslash}m{1cm} >{\centering\arraybackslash}m{1cm} >{\centering\arraybackslash}m{1cm}}
        \Xhline{1pt}
        Architecture & IoU(\%) & F1(\%) & P(\%) & R(\%) \\ \hline
        Parallel \cite{DBLP:journals/tgrs/WangFM022}  & 81.44 & 89.77 & 90.75 & 88.81  \\
        Sequential \cite{DBLP:journals/corr/abs-2102-04306} & 82.84 & 90.62 & 91.60 & 89.66 \\
        Alternative \cite{DBLP:journals/lgrs/ZhangWZZ23} & 82.15 & 90.20 & 91.50 & 88.94 \\
        Ours & \textbf{83.74} & \textbf{91.15} & \textbf{92.09} & \textbf{90.22} \\ 
        \Xhline{1pt}
    \end{tabular}
    \vspace{-10pt}
  \end{center}
\end{table}

\noindent\textbf{Ablation Study of Encoder Network.} We conduct an ablation study to verify the effectiveness of the proposed cooperative encoder. Table \ref{Comparison:Hybrid} shows the quantitative comparison of different hybrid architectures on Inria aerial dataset. We refer to the encoder networks in Buildformer \cite{DBLP:journals/tgrs/WangFM022}, TransUNet \cite{DBLP:journals/corr/abs-2102-04306} and DSATNet \cite{DBLP:journals/lgrs/ZhangWZZ23} as parallel, sequential and alternative variants of the hybrid CNN and transformer blocks, respectively. Compared to the three existing hybrid architectures for building extraction, our cooperative encoder achieves the improvements of $2.30\%$, $0.90\%$ and $1.59\%$ in the IoU metric. The impressive performance gains indicate that our cooperative encoder can effectively narrow the inherent gap of the feature pyramids.

\begin{table}[h]
  \begin{center}
    \caption{Ablation Experimental Results Of Our Decoder On Inria Aerial Image Labeling Dataset}
    \renewcommand\arraystretch{1.4}
    \label{Ablation:Decoder}
    \begin{tabular}{>{\centering}m{2.2cm}|>{\centering\arraybackslash}m{1cm} >{\centering\arraybackslash}m{1cm} >{\centering\arraybackslash}m{1cm} >{\centering\arraybackslash}m{1cm}}
        \Xhline{1pt}
        Decoder & IoU(\%) & F1(\%) & P(\%) & R(\%) \\ \hline
        FPN \cite{DBLP:conf/cvpr/LinDGHHB17} & 82.78 & 90.58 & 91.32 & 89.85 \\
        UperHead \cite{DBLP:conf/eccv/XiaoLZJS18} & 82.83 & 90.61 & 91.30 & 89.92  \\
        ASPPHead \cite{DBLP:conf/eccv/ChenZPSA18} & 83.26 & 90.86 & 91.67 & 90.07 \\
        Ours & \textbf{83.74} & \textbf{91.15} & \textbf{92.09} & \textbf{90.22} \\ 
        \Xhline{1pt}
    \end{tabular}
    \vspace{-10pt}
  \end{center}
\end{table}

\noindent\textbf{Ablation Study of Decoder Network.} Table \ref{Ablation:Decoder} shows the quantitative comparison of different decoders on the Inria aerial dataset. We take FPN \cite{DBLP:conf/cvpr/LinDGHHB17}, UperHead \cite{DBLP:conf/eccv/XiaoLZJS18} and ASPPHead \cite{DBLP:conf/eccv/ChenZPSA18} as comparative variants to validate the effectiveness of our decoder. The variant method using FPN obtains the IoU and F1 scores of $82.78\%$ and $90.58\%$. The ASPPHead obtains higher performance due to the larger reception field perceived by the atrous spatial pyramid pooling operation. Our UAGLNet using UAD achieves more competitive performance with the corresponding scores of $83.74\%$ and $91.15\%$, which obtains improvement of $0.96\%$ and $0.57\%$ compared to FPN.

\begin{table}[h]
  \begin{center}
    \caption{Ablation Study of Encoder Global-Local Balance and Decoder Global-Local Fusion}
    \label{Ablation:CIB}
    \renewcommand\arraystretch{1.4}
    \begin{tabular}{ >{\centering}m{1.5cm}  >{\centering\arraybackslash}m{1.5cm} | >{\centering\arraybackslash}m{0.7cm}  >{\centering\arraybackslash}m{0.7cm}  >{\centering\arraybackslash}m{0.7cm} >{\centering\arraybackslash}m{0.7cm}}
    \Xhline{1pt}
    Global-Local Interaction & Global-Local Fusion & IoU(\%) & F1(\%) & P(\%) & R(\%) \\ \hline
    \XSolidBrush & \XSolidBrush & 82.02 & 90.22 & 91.42 & 88.86 \\
    CIB & \XSolidBrush & 82.43 & 90.37 & 91.18 & 89.57  \\
    \XSolidBrush & GLF & 83.08 & 90.76 & 91.59 & 89.94 \\
    CIB & GLF & \textbf{83.33} & \textbf{90.90} & \textbf{91.85} & \textbf{89.97} \\ 
    \Xhline{1pt}
    \end{tabular}
    \vspace{-10pt}
  \end{center}
\end{table}

\noindent\textbf{Ablation Study of CIB and GLF.} In UAGLNet, we introduce the Cooperative Interaction Block (CIB) to balance the global and local representations. To verify the effectiveness of the global-local balance in the encoder, we conduct an ablation study by selectively incorporating CIB and GLF for comparison. As shown in Table \ref{Ablation:CIB}, the results demonstrate that CIB and GLF can effectively improve the performance in all of the metrics.

\begin{table}[!t]
  \begin{center}
    \caption{Ablation Study of hyperparameters setting in MKFM on Inria Aerial Image Labeling Dataset}
    \label{Ablation:hyperparameter}
    \renewcommand\arraystretch{1.4}
    \begin{tabular}{ >{\centering\arraybackslash}m{2.5cm} | >{\centering\arraybackslash}m{1cm}  >{\centering\arraybackslash}m{1cm}
    >{\centering\arraybackslash}m{1cm} >{\centering\arraybackslash}m{1cm}}
    \Xhline{1pt}
    Hyperparameters & IoU(\%) & F1(\%) & P(\%)  & R(\%) \\ \hline
    $k = 3, n = 1$ & 83.12 & 90.78 & 91.86 & 89.73 \\
    $k = 5, n = 2$ & 83.44 & 90.97 & 91.80 & 90.16 \\
    $k = 9, n = 4$ & \textbf{83.74} & \textbf{91.15} & \textbf{92.09} & \textbf{90.22}\\
    $k = 17, n = 8$ & 83.31 & 90.89 & 91.86 & 89.95 \\ 
    \Xhline{1pt}
    \end{tabular}
    \vspace{-15pt}
  \end{center}
\end{table}

\noindent\textbf{Hyperparameter Setting in MKFM.} In MKFM, we utilize a multi-kernel structure to enhance the model's perceptional capacity while maintaining a controllable computational budget. It's important to note that $n$ must be divisible by the feature dimension (64), and $k$ is defined as $k=2n+1$. To evaluate the effectiveness of these hyperparameter settings, we conducted an ablation study, and the results are presented in Table \ref{Ablation:hyperparameter}. The results show that setting $n$ to 4 and $k$ to 9 yields the best performance for our model.

\begin{table}[H]
  \begin{center}
    \caption{Ablation Experimental Results Of Different Global-Local Fusion Strategy On The Inria Aerial Image Labeling Dataset}
    \renewcommand\arraystretch{1.4}
    \label{Ablation:Fusion}
    \begin{tabular}{c c| c c c c }
    \Xhline{1pt}
        $\mathbf{F}_{L}$ & $\mathbf{F}_{G}$ & IoU(\%) & F1(\%) & P(\%) & R(\%) \\ \hline
        $\{ \mathbf{F}_{1},\mathbf{F}_{2}\}$ & $\{ \mathbf{F}_{4} \}$  & 83.13 & 90.79 & 91.66 & 89.94  \\
        $\{ \mathbf{F}_{1} , \mathbf{F}_{2} , \mathbf{F}_{3} \}$  & $\{ \mathbf{F}_{4} \}$  & 83.36 & 90.92 & 91.52 & \textbf{90.33} \\
        $\{ \mathbf{F}_{1} , \mathbf{F}_{2} \}$  & $\{ \mathbf{F}_{3} , \mathbf{F}_{4} \}$  & 83.52 & 91.02 & 91.85 & 90.21  \\
        $\{ \mathbf{F}_{1} , \mathbf{F}_{2} , \mathbf{F}_{3} \}$  & $\{ \mathbf{F}_{3} , \mathbf{F}_{4} \}$  & \textbf{83.74} & \textbf{91.15} & \textbf{92.09} & 90.22 \\ 
    \Xhline{1pt}
    \end{tabular}
        \vspace{-15pt}
  \end{center}
\end{table}


\noindent\textbf{Ablation Study of Feature Fusion Strategy.} To verify the effectiveness of our global-local fusion strategy within GLF, we conduct an ablation study to test the variants with different fusion strategies. The ablation results are shown in Table \ref{Ablation:Fusion}. If we select $\{\mathbf{F}_{1}, \mathbf{F}_{2} \}$ to construct local representation $\mathbf{F}_{L}$ and $\{\mathbf{F}_{4} \}$ for global representation $\mathbf{F}_{G}$, the variant yields $83.13\%$ in the IoU metric. By incorporating $\mathbf{F}_{3}$ into the generation of $\mathbf{F}_{L}$, the variant obtains the IoU performance gain of $0.23\%$. Since local context is enhanced by $\mathbf{F}_{3}$, this variant is capable to discover more foreground pixels but also exhibits more misclassifications, resulting in higher recall and lower precision. Meanwhile, by incorporating $\mathbf{F}_{3}$ into the global representation $\mathbf{F}_{G}$, the variant obtains the IoU score improvement of $0.39\%$. If $\mathbf{F}_{3}$ is utilized for both $\mathbf{F}_{L}$ and $\mathbf{F}_{G}$, the variant can achieve the best performance with the IoU score of $83.74\%$. 



\subsection{Discuss on the Computation Efficiency}
We discuss the computational efficiency of our UAGLNet by measuring the frames per second (FPS) and parameters (Params). We take the CNN-based model HRNet \cite{DBLP:conf/cvpr/0009XLW19}, the recent state-of-the-art transformer-based models such as SDSCUNet \cite{SDSCunet} and UANet\_PVT \cite{DBLP:journals/tgrs/LiHCZZ24} for comparison, the results presented in Table \ref{Results:efficiency}. The experiment shows that UAGLNet achieves the highest IoU and F1 scores with 27.53 fps, which is $42.05\%$ faster than SDSCUNet. Besides, it only requires $15.34$M, which reduce $28.05\%$ parameters compared to SDSCUNet.

\begin{table}[!t]
  \begin{center}
    \caption{Comparison of State-of-the-Art Methods and Ours for Real-Time Applications}
    \label{Results:efficiency}
    \renewcommand\arraystretch{1.4}
    \begin{tabular}{ >{\centering\arraybackslash}m{2.2cm} | >{\centering\arraybackslash}m{1cm}  >{\centering\arraybackslash}m{1cm}
    >{\centering\arraybackslash}m{1cm} >{\centering\arraybackslash}m{1.2cm}
    >{\centering\arraybackslash}m{1.2cm}}
    \Xhline{1pt}
    Methods & IoU(\%) & F1(\%) & FPS & Params  \\ \hline
    HRNet \cite{DBLP:conf/cvpr/0009XLW19} & 77.14 & 87.10 & 19.59 & 67.17M  \\
    SDSCUNet \cite{SDSCunet} & 83.01& 90.71 & 19.38 & 21.32M  \\
    UANet\_PVT \cite{DBLP:journals/tgrs/LiHCZZ24} & 83.34 & 90.91  & 23.89 & 25.64M \\
    UAGLNet (ours) & \textbf{83.74} & \textbf{91.15} & \textbf{27.53} & \textbf{15.34M}  \\ 
    \Xhline{1pt}
    \end{tabular}
    \vspace{-15pt}
  \end{center}
\end{table}

\begin{table}[h]
  \begin{center}
    \caption{Comparative analyses of High-resolution Prediction Results On Massachusetts Building Dataset.}
    \label{Comparison:Mass_2}
    \renewcommand\arraystretch{1.4}
    \begin{tabular}{p{1.4cm}<{\centering}p{0.6cm}<{\centering}p{0.7cm}<{\centering}p{0.7cm}<{\centering}p{0.7cm}<{\centering}p{0.7cm}<{\centering}p{1cm}<{\centering}}
    \Xhline{1pt}
        Method & Year & IoU(\%) & F1(\%) & P(\%) & R(\%) & FLOPs \\ \hline
        UANet \cite{DBLP:journals/tgrs/LiHCZZ24} & 2024 & 70.75 & 82.87 & 86.21 & 79.78 & 1004.75G \\ 
        UAGLNet & & \textbf{72.32} & \textbf{83.94} & \textbf{87.08} & \textbf{81.01} & \textbf{419.58G} \\
    \Xhline{1pt}
    \end{tabular}
    \vspace{-15pt}
  \end{center}
\end{table}

\subsection{Discuss on High-Resolution Images}
We test UAGLNet and UANet on the high-resolution Massachusetts Building dataset without image cropping, the size of each image is $1500 \times 1500$. The results are presented in Table \ref{Comparison:Mass_2}, we can see that for the high-resolution input setting, UAGLNet obtains obvious performance gains, it only requires less than half computational flop compared to UANet. This is because UAGLNet integrates multi-kernel feature modulator(MKFM) with lightweight depth-wise separable convolution and transformer blocks, which significantly reduce the computational complexity.

\noindent\textbf{Discuss on UAD's  Robustness.}
To evaluate the effectiveness of UAGLNet in more challenging scenarios, we synthesize the low-resolution and noisy images using the Inria Aerial Image Labeling dataset. Specifically, the low resolution version degrades the resolution by a factor of $16$, the noisy version adds Gaussian noise to the images with the standard deviation $5$. The comparative results are demonstrated in Table \ref{Ablation:noise}. We can see that compared to the variant approach without Uncertainty-Aggregated Decoder (UAD), our UAGLNet obtains higher performance in all metrics, indicating that UAD is more robust if the dataset is contaminated by the low-resolution or noisy images. 

\begin{table}[ht!]
  \begin{center}
    \caption{Ablation experimental results of our UAD on the synthesis low-resolution and noise variants of Inria Aerial Image Labeling Dataset.}
    \renewcommand\arraystretch{1.4}
    \label{Ablation:noise}
    \begin{tabular}{>{\centering}m{2.8cm}|>{\centering\arraybackslash}m{1cm} >{\centering\arraybackslash}m{1cm} >{\centering\arraybackslash}m{1cm} >{\centering\arraybackslash}m{1cm}}
    \Xhline{1pt}
        Low Resolution & IoU(\%) & F1(\%) & P(\%) & R(\%) \\ \hline
        w/o UAD & 80.32 & 89.09 & 89.74 & 88.45  \\
        + UAD (ours) & 81.19 & 89.62 & 90.64 & 88.62 \\ \hline \hline
        Add Noise & IoU(\%) & F1(\%) & P(\%) & R(\%) \\ \hline
        w/o UAD & 82.72 & 90.54 & 91.35 & 89.75 \\
        + UAD (ours) & 83.41 & 90.96 & 91.91 & 90.02 \\  
    \Xhline{1pt}
    \end{tabular}
    \vspace{-15pt}
  \end{center}
\end{table}

\subsection{Feature Visualizations}

\begin{figure*}[ht!]
  \centering
  \includegraphics[width=6.4in]{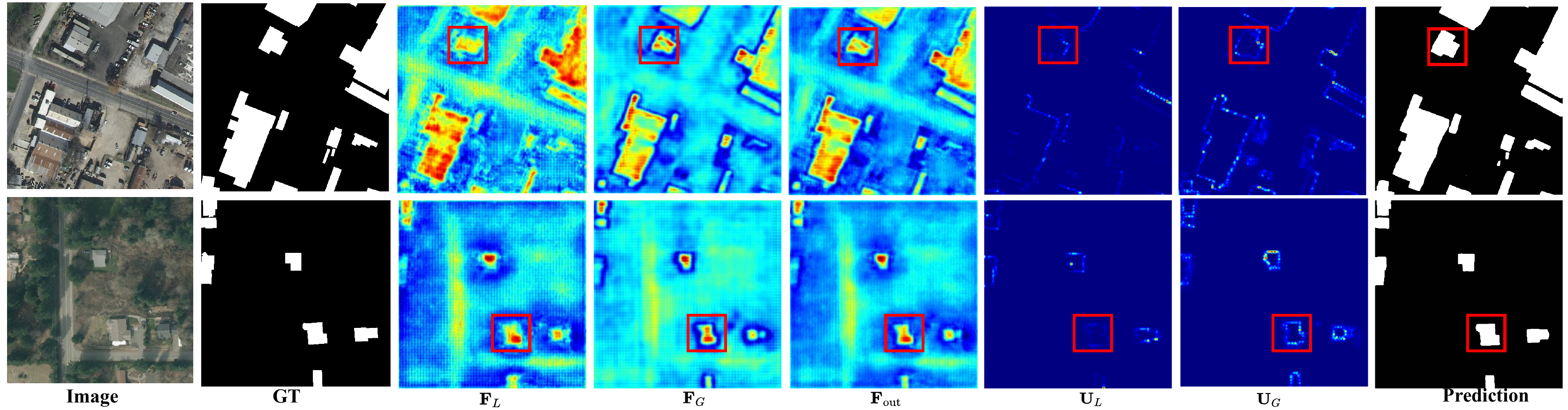}
  \caption{The visualized features in the Uncertainty-Aggregated Decoder (UAD). The irrelevant noises in $\mathbf{F}_{L}$ are suppressed and the  low-confidence uncertain regions in $\mathbf{F}_{G}$ are aggregated within the red rectangles. }
  \label{uncertainty_add1}
  \vspace{-15pt}
\end{figure*}

We add more examples to demonstrate the effectiveness of the uncertainty prediction in UAGLNet. As shown Fig. \ref{uncertainty_add1}, the local feature $\mathbf{F}_{L}$ within the red box region contains more details, while the global feature $\mathbf{F}_{G}$ is more discriminative. To obtain accurate prediction, the local uncertainty $\mathbf{U}_{L}$ pays more attention to filter out the irrelevant details and the global uncertainty $\mathbf{U}_{G}$ focuses on maintaining reliable boundaries, leading the fused output $\mathbf{F}_{\mathrm{out}}$ to be more stable against the uncertain noises. 

\subsection{Generalization Analyse of UAGLNet}
We add a cross-dataset experiment to assess UAGLNet's generalization capability. Here we train all of the methods on Inria dataset and test them on WHU dataset for fair comparison.  As shown in Table \ref{Ablation:OutOfDomain}, the performance of the CNN based method like HRNet drops  $15.69\%$ and $9.65\%$ in terms of IoU and F1, respectively. The transformer based approach BuildFormer drops  $15.88\%$ and $9.46\%$. Compared to recent state-of-the-art methods, UAGLNet exhibits superior generalization capability with $7.87\%$ and $4.45\%$ performance degradation.
\begin{table}[!t]
  \begin{center}
    \caption{Comparison Of The State-Of-The-Art Methods And Ours in Cross-Domain Testing On The WHU Building Dataset .}
    \renewcommand\arraystretch{1.2}
    \label{Ablation:OutOfDomain}
    \begin{tabular}{ >{\centering\arraybackslash}m{2.1cm} | >{\centering\arraybackslash}m{0.7cm}  >{\centering\arraybackslash}m{0.7cm}
    >{\centering\arraybackslash}m{1cm} >{\centering\arraybackslash}m{1cm}}
    \Xhline{1pt}
        \multirow{2}{*}{Methods} & \multicolumn{2}{c|}{Trained on WHU} & \multicolumn{2}{c}{Trained on Inria} \\ \cline{2-5}
         & IoU(\%) & F1(\%) & IoU(\%) & F1(\%) \\ \hline
        UNet \cite{DBLP:conf/miccai/RonnebergerFB15} & 88.08 & 93.66 & 67.06 (21.02$\downarrow$) & 80.28 (13.38$\downarrow$) \\
        HRNet \cite{DBLP:conf/cvpr/0009XLW19} & 88.21 & 93.73 & 72.52 (15.69$\downarrow$) & 84.08 (9.65$\downarrow$)  \\
        BuildFormer \cite{DBLP:journals/tgrs/WangFM022} & 91.44 & 95.53 & 75.56 (15.88$\downarrow$) & 86.07 (9.46$\downarrow$) \\
        UAGLNet (ours) & \textbf{92.07} & \textbf{95.87} & \textbf{84.20} (\textbf{7.87$\downarrow$}) & \textbf{91.42} (\textbf{4.45$\downarrow$}) \\ 
    \Xhline{1pt}
    \end{tabular}
    \vspace{-15pt}
  \end{center}
\end{table}

\subsection{Future Work}
We have noticed that the proposed UAGLNet can be future applied to other remote sensing task satellite imagery. For example, it can be rapidly deployed to other satellite imagery based semantic segmentation datasets by simply replacing the decoder head for task-specific model training. While UAGLNet can already run at real-time, the running speed can be  further improved by introducing model lightweight tricks such as distillation, pruning. Besides, we would explore to integrate UAGLNet and super-resolution module into a unified framework for the challenging low-resolution images.

\section{Conclusions}
In this paper, we propose an Uncertainty-Aggregated Global-Local Fusion Network (UAGLNet) to exploit high-quality global-local visual semantics under the guidance of uncertainty modeling. We first design a cooperative encoder, which adopts convolutional and non-local self-attention operators at different stages to capture the local and global visual semantics. An intermediate cooperative interaction block (CIB) is designed to narrow the gap between the local and global semantics when the network becomes deeper. Afterwards, we propose a Global-Local Fusion (GLF) module to complementarily fuse the hierarchical visual representations. Moreover, we propose an Uncertainty-Aggregated Decoder (UAD) to mitigate the segmentation uncertainty in ambiguous regions. Extensive experiments demonstrate that our method achieves superior performance on the public benchmarks.

\bibliographystyle{IEEEtran}
\bibliography{document}

\vspace{-15mm}
\begin{IEEEbiography}[{\includegraphics[width=0.8in]{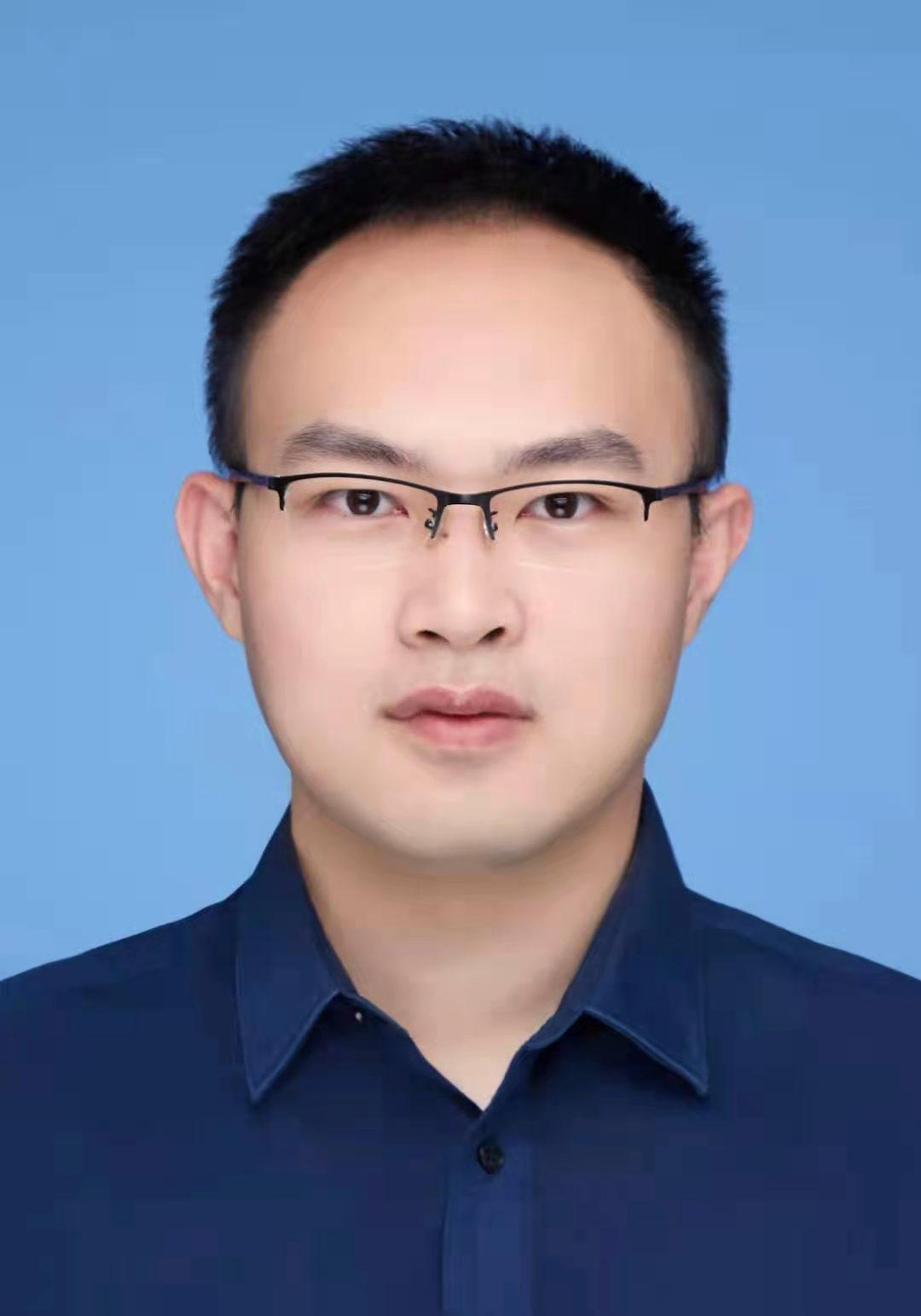}}]{Siyuan Yao} received the Ph.D. degree from Institute of Information Engineering, Chinese Academy of Sciences, in 2022. He is currently an Assistant Professor with the School of Computer Science, Beijing University of Posts and Telecommunications (BUPT), China. He was supported by the Tencent Rhino-Bird Elite Talent Training Program in 2021. His research interests include visual object tracking, video/image analysis and machine learning.
\end{IEEEbiography}
\vspace{-15mm}
\begin{IEEEbiography}[{\includegraphics[width=0.8in]{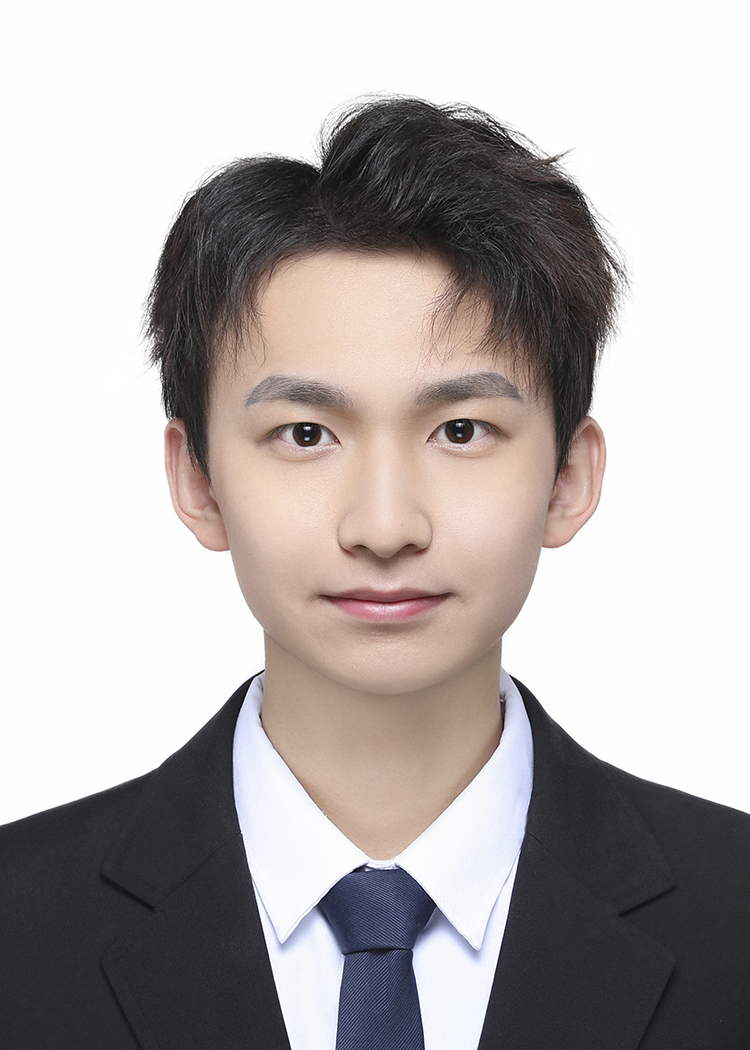}}]{Dongxiu Liu} received the B.E. degree from Beijing University of Posts and Telecommunications, China, where he is currently pursuing M.S. degree in computer science. His research interests include image segmentation, scene understanding, and spatial reasoning.
\end{IEEEbiography}
\vspace{-15mm}
\begin{IEEEbiography}[{\includegraphics[width=0.8in]{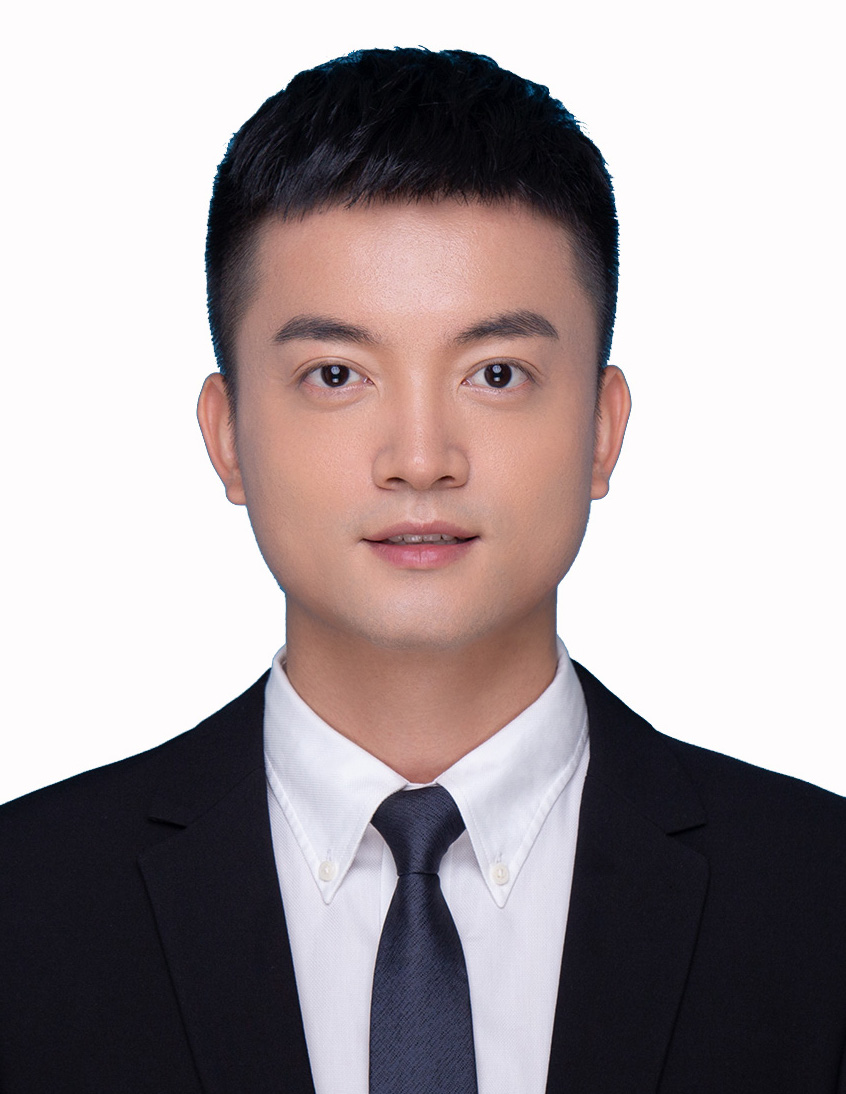}}]{Taotao Li} received the Ph.D. degree in cyber security from Institute of Information Engineering, Chinese Academy of Sciences and University of Chinese Academy of Sciences, China, in 2022. He has worked as a Research Fellow with Sun Yat-Sen University, China. He is currently an assistant professor with the School of Software Engineering, Sun Yat-Sen University, China. His main research interests include blockchain, Data security, and applied cryptography.
\end{IEEEbiography}
\vspace{-15mm}

\begin{IEEEbiography}[{\includegraphics[width=0.8in]{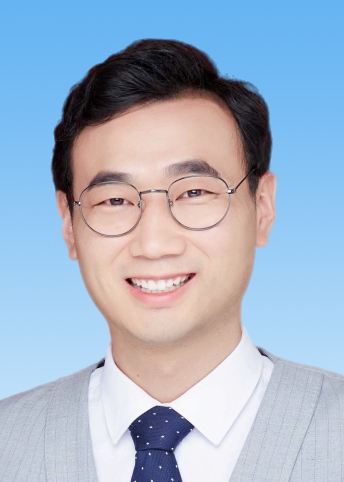}}] {Shengjie Li} received the Ph.D. degree in information and communication engineering from the Beijing University of Posts and Telecommunications (supervisor: Prof. Junliang Chen) in 2020. He is currently a Lecturer with the State Key Laboratory of Networking and Switching Technology at Beijing University of Posts and Telecommunications. His current research interests include Internet of Things technology and visual object tracking.
\end{IEEEbiography}
\vspace{-15mm}

\begin{IEEEbiography}[{\includegraphics[width=0.8in]{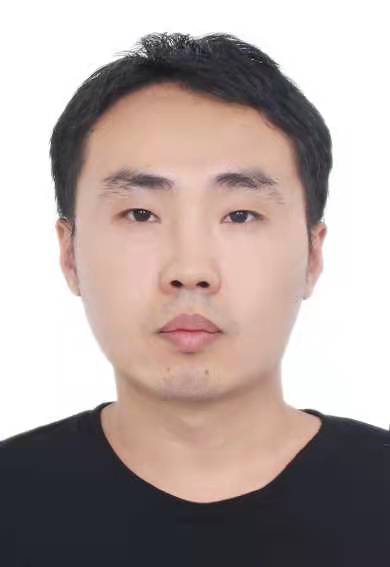}}]{Wenqi Ren} received the Ph.D.
degree from Tianjin University, Tianjin, China, in 2017. From 2015 to 2016, he was supported by China Scholarship Council and working with
Prof. Ming-Husan Yang as a Joint-Training Ph.D. Student with the Electrical Engineering and Computer Science Department, University of California at Merced. He is currently a Professor with the School of Cyber Science and Technology, Sun Yatsen University, Shenzhen Campus, Shenzhen, China. His research interests include image processing and related high-level vision problems. He received the Tencent Rhino Bird Elite Graduate Program Scholarship in 2017 and the MSRA Star Track Program in 2018.
\end{IEEEbiography}
\vspace{-15mm}
\begin{IEEEbiography}[{\includegraphics[width=0.8in]{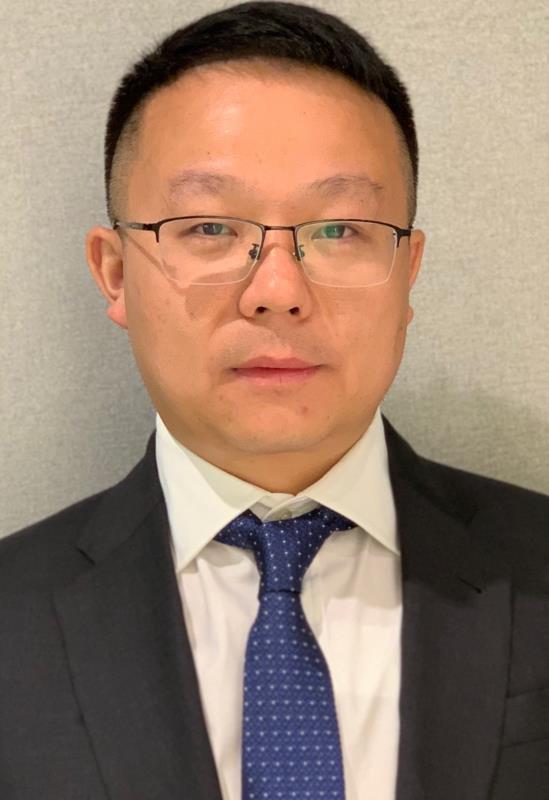}}]{Xiaochun Cao} received the BE and ME degrees in computer science from Beihang University (BUAA), China, and the PhD degree in computer science from the University of Central Florida, USA. He is a professor and dean with the School of Cyber Science and Technology, Shenzhen Campus of Sun Yat-sen University. His dissertation nominated for the university level Outstanding Dissertation Award. After graduation, he spent about three years with ObjectVideo Inc. as a research scientist. From 2008 to 2012, he was a professor with Tianjin University. Before joining SYSU, he was a professor with the Institute of Information Engineering, Chinese Academy of Sciences. He has authored and coauthored more than 200 journal and conference papers. In 2004 and 2010, he was the recipients of the Piero Zamperoni best student paper award at the International Conference on Pattern Recognition. He is on the editorial boards of IEEE Transactions on Pattern Analysis and Machine Intelligence and IEEE Transactions on Image Processing, and was on the editorial boards of IEEE Transactions on Circuits and Systems for Video Technology and IEEE Transactions on Multimedia.
\end{IEEEbiography}

\end{document}